\documentclass[final,5p,times,twocolumn]{elsarticle}

\usepackage{graphicx}
\usepackage{subcaption}
\usepackage{hyperref}
\usepackage{rotating}
\usepackage{booktabs}
\usepackage{multirow}
\usepackage{amsfonts}
\usepackage{xcolor}
\usepackage{bm}
\usepackage{mathtools}
\usepackage{amsmath}
\usepackage{amssymb}
\usepackage{amsthm}
\usepackage{enumitem}
\usepackage{tabularx}
\usepackage{colortbl}
\usepackage{makecell}

\journal{Future Generation Computer Systems}

\begin{document}

\begin{frontmatter}

%% ----------------------
%% Title
%% ----------------------
\title{FedVAR: Prototype-Aligned Federated Framework for Video Anomaly Recognition}

%% ----------------------
%% Authors
%% ----------------------
\author[1]{Ghani Haider}
\ead{g.haider@cbnu.ac.kr}

\author[1]{Majid Kundroo}
\ead{kundroomajid@cbnu.ac.kr}

\author[2]{Boyun Eom}
\ead{eby@etri.re.kr}

\author[2]{Dong-Hwan Park\corref{cor1}}
\ead{dhpark@etri.re.kr}

\author[3]{Chen Chen}
\ead{chen.chen@ucf.edu}

\author[1]{Taehong Kim\corref{cor1}}
\ead{taehongkim@cbnu.ac.kr}

%% ----------------------
%% Corresponding Authors
%% ----------------------
\cortext[cor1]{Corresponding authors.}

%% ----------------------
%% Affiliations
%% ----------------------

\affiliation[1]{
    organization={School of Information and Communication Engineering, Chungbuk National University},
    city={Cheongju},
    postcode={28644},
    country={Republic of Korea}
}

\affiliation[2]{
    organization={Electronics and Telecommunications Research Institute (ETRI)},
    city={Daejeon},
    postcode={34129},
    country={Republic of Korea}
}

\affiliation[3]{
    organization={Institute of Artificial Intelligence (IAI), University of Central Florida},
    city={Orlando},
    postcode={32816},
    country={USA}
}

\begin{abstract}
In the era of Industrial Internet of Things (IIoT) and Cyber-Physical Systems (CPS), Federated Learning (FL) offers a promising decentralized intelligence paradigm for Video Anomaly Recognition (VAR). This task is vital for maintaining high-fidelity Digital Twins and ensuring safety in mission-critical environments. However, the inherent data heterogeneity across distributed edge clients leads to a fundamental challenge known as semantic misalignment, where clients learn divergent feature representations of ``normal'' and ``abnormal'' events. The problem becomes particularly pronounced in VAR, where the presence of diverse and fine-grained anomaly categories leads each client to develop distinct semantic interpretations of abnormality. Existing federated methods primarily focus on binary anomaly detection and fail to address this misalignment, preventing effective fine-grained recognition. In this paper, we introduce FedVAR, a weakly-supervised FL framework explicitly designed for VAR. Leveraging the rich representations of Vision-Language Models (VLMs), FedVAR employs a prototype-based alignment mechanism that creates a shared semantic anchor for all clients to re-center and align their visual and textual feature spaces. This process enforces a consistent representation of ``normality'' across the decentralized network, directly mitigating semantic misalignment and enabling robust prompt-learning of anomaly direction vectors with minimal communication overhead. We conduct extensive experiments on challenging benchmarks under various non-IID data partitioning schemes, unseen domains, and novel anomaly classes. The results demonstrate that FedVAR consistently outperforms state-of-the-art federated baselines, establishing a robust framework for distributed intelligence in video-based CPS.
\end{abstract}


\end{frontmatter}

\def\thefootnote{}\footnotetext{\textcopyright~2026. This manuscript version is made available under the CC-BY-NC-ND 4.0 license (\url{http://creativecommons.org/licenses/by-nc-nd/4.0/}). This is the accepted manuscript version of an article published in \textit{Future Generation Computer Systems}. The final published version is available at \url{https://doi.org/10.1016/j.future.2026.108745}.}
\def\thefootnote{\arabic{footnote}}

\section{Introduction}\label{sec:intro}
The task of identifying unusual events in video streams, known as Video Anomaly Detection (VAD) \cite{abdalla2025video}, has become a cornerstone of automated surveillance \cite{ullah2022artificial} and industrial monitoring within complex Cyber-Physical Systems (CPS) \cite{politi2025real, bedogni2025fluid}. However, merely detecting an anomaly is often insufficient for mission-critical applications \cite{pathirannahalage2025comprehensive}. A more advanced and challenging task is Video Anomaly Recognition (VAR) \cite{hussain2026class}, which aims to jointly detect the presence of an anomaly and identify its specific semantic category, such as human activities (e.g., “fighting”, “vandalism”) or industrial hazard events (e.g., “equipment malfunction” or “fire”). This fine-grained understanding is critical for triggering appropriate, event-specific responses in domains ranging from smart energy systems to urban mobility \cite{alem2023novel, djebali2024survey, elshenawy2018towards}. Despite its importance, VAR remains exceptionally challenging due to the rare, diverse, and ambiguous nature of anomalies, coupled with the extreme data imbalance between normal and abnormal events \cite{ucfcrime_sultani2018real}.

To tackle these challenges without relying on prohibitively expensive frame-level annotations, recent advancements have leveraged Weakly Supervised Video Anomaly Detection (WSVAD) \cite{karim2024real} alongside powerful Vision-Language Models (VLMs) like CLIP \cite{zsclip_icml2021}. Approaches such as AnomalyCLIP \cite{anomalyclip} have demonstrated that aligning visual features with textual anomaly descriptors can enable remarkable fine-grained recognition. However, these paradigms fundamentally assume centralized data access. In real-world deployments, such as urban surveillance networks, centralizing video footage is often untenable due to strict privacy regulations, data sovereignty policies, and bandwidth constraints \cite{salim2024digital, barbone2025device}. 

Federated Learning (FL) \cite{mcmahan2017communication} provides a promising privacy-preserving alternative by allowing edge clients to collaboratively train a shared model without exchanging raw data. While FL has been successfully applied to binary VAD, its extension to fine-grained VAR remains unexplored due to the critical challenge of semantic misalignment. In a federated surveillance network, each client captures a highly distinct environment and observes different subsets of anomalies due to non-Independent and Identically Distributed (non-IID) class distributions. 
For example, in a smart-city surveillance camera network, a highway camera may predominantly capture traffic ``accidents'', whereas a retail camera may primarily observe ``shoplifting'', exposing clients to highly heterogeneous environmental contexts and anomaly distributions.
Consequently, each client learns a distinct statistical representation of what constitutes ``normal'' and ``abnormal'' behavior. Existing federated VAD frameworks, such as Fed-WSVAD \cite{fedwsvad_aaai2025} and CLAP \cite{clap_cvpr2024}, focus purely on binary detection and do not explicitly align these heterogeneous feature spaces. When extended to multi-class VAR, this architectural gap causes the global model's semantic representations to collapse, rendering naive model averaging ineffective. 

To overcome this, we propose FedVAR, a federated framework specifically designed for distributed intelligence in video anomaly recognition. Our core contribution lies in a novel prototype-based alignment mechanism that bridges the semantic gap between heterogeneous clients. FedVAR leverages CLIP's robust visual-textual representations as a foundation. Rather than allowing clients' local feature spaces to diverge, FedVAR computes local normality prototypes at the edge and aggregates them on the central server to construct a single global normality prototype. This global prototype acts as a shared semantic anchor. By re-centering their local visual and textual features around this shared anchor, all participating clients evaluate anomalies from an aligned semantic origin. This directly mitigates the semantic misalignment caused by non-IID data, enabling the global model to learn robust, generalizable, and fine-grained anomaly representations while preserving strict data privacy.
Our main contributions are summarized as follows:
\begin{itemize}
    \item We propose FedVAR, the first framework designed specifically to address the task of fine-grained video anomaly recognition in a privacy-preserving federated learning setting, facilitating decentralized intelligence.
    \item We introduce a novel cross-modal prototype alignment mechanism that constructs a global normality prototype. This serves as a shared semantic anchor, effectively mitigating feature space misalignment caused by non-IID client data.
    \item We conduct extensive experiments on multiple benchmarks, including cross-domain transfer and unseen anomaly class generalization. We demonstrate that FedVAR consistently outperforms existing federated anomaly detection frameworks across both detection and recognition metrics, establishing a strong new baseline for federated VAR.
\end{itemize}

\section{Related Work}\label{sec:relatedwork}
\begin{table*}[t]
\centering
\caption{Comparison of the proposed FedVAR with state-of-the-art baselines. We highlight key mechanisms, advantages, and limitations, specifically emphasizing robustness to non-IID data and semantic misalignment.}
\label{tab:related_work}
\renewcommand{\arraystretch}{1.2}
\small
\begin{tabularx}{\textwidth}{@{} 
    >{\raggedright\arraybackslash}p{2.3cm} 
    >{\raggedright\arraybackslash}p{1.7cm} 
    >{\raggedright\arraybackslash}X 
    >{\raggedright\arraybackslash}X 
    >{\raggedright\arraybackslash}X @{}}
\toprule
\textbf{Method} & \textbf{Task} & \textbf{Key Mechanism} & \textbf{Advantages} & \textbf{Limitations} \\
\midrule

\textbf{ZS-CLIP} \cite{zsclip_icml2021} \newline (ICML 2021) & 
Centralized Image Classification & 
Contrastive VLM pre-training with manual text prompts. & 
Strong zero-shot open-vocabulary representations. & 
No temporal modeling; rigid prompts; requires centralized data. \\
\midrule

\textbf{Temp-CLIP} \cite{temporalclip_eccv2022} \newline (ECCV 2022) & 
Centralized Video Understanding & 
VLM adaptation via temporal transformer \& learnable prompts. & 
Parameter-efficient; captures dynamic temporal correlations. & 
Requires centralized data; cannot handle non-IID environments. \\
\midrule

\textbf{PPVU} \cite{ppvu_dsc2023} \newline (DSC 2023) & 
Federated Video Understanding & 
FL framework utilizing a TimeSformer backbone. & 
Privacy-preserving spatiotemporal feature extraction. & 
High communication overhead (full model updates); lacks VLM guidance. \\
\midrule

\textbf{FedCoOp} \cite{fedcoop_tmc2023} \newline (TMC 2023) & 
Federated Image Classification & 
Federated continuous prompt learning (frozen VLM backbone). & 
Communication-efficient; preserves privacy via prompt-only updates. & 
Assumes homogeneous labels; no temporal modeling; vulnerable to client drift. \\
\midrule

\textbf{CLAP} \cite{clap_cvpr2024} \newline (CVPR 2024) & 
Federated VAD & 
Unsupervised FL via GMM clustering \& pseudo-labeling. & 
Fully unsupervised; eliminates expensive annotation costs. & 
Inconsistent pseudo-labels on non-IID data; limited to binary VAD. \\
\midrule

\textbf{Fed-WSVAD} \cite{fedwsvad_aaai2025} \newline (AAAI 2025) & 
Federated WSVAD & 
Multimodal prompt generation via global/local contexts. & 
Effective binary anomaly localization; balances personalization. & 
Lacks cross-client alignment; suffers from semantic misalignment in extreme non-IID settings. \\
\midrule

\rowcolor{gray!10} 
\textbf{FedVAR} \newline \textbf{(Ours)} & 
Federated VAR & 
Cross-modal global prototype alignment \& axial temporal attention. & 
Resolves semantic misalignment via shared anchor; robust to non-IID; enables multi-class VAR. & 
Coarse temporal aggregation may dilute boundaries in highly sparse videos. \\
\bottomrule
\end{tabularx}
\end{table*}

\subsection{Video Anomaly Detection under Centralized and Federated Learning}\label{subsec:relatedwork_vad}
Centralized VAD methods, which assume all training data are located on a single server, have evolved significantly. Early approaches focused on reconstruction-based criteria, using models like autoencoders~\cite{gong2019memorizing} or predictive models~\cite{liu2018future} to learn the distribution of normal events and detect deviations. To capture more complex temporal dynamics, subsequent works integrated powerful deep learning architectures, including 3D CNNs~\cite{qasim2023video}, graph convolutional networks for modeling object interactions~\cite{chen2023multiscale}, and transformers for long-range dependency modeling~\cite{biradar2024robust}. To reduce annotation costs, weakly supervised methods~\cite{ucfcrime_sultani2018real, wu2024open, lv2023unbiased} became prominent, leveraging Multiple Instance Learning (MIL) with only video-level labels. Despite their progress in modeling complexity, these methods share a fundamental limitation of relying on a centralized data repository, which is often impractical due to privacy concerns and data governance policies~\cite{lu2024federated}.

FL offers a privacy-preserving alternative by training on decentralized data~\cite{KUNDROO2026103843}. While FL has been explored for anomaly detection in tabular and image data~\cite{xiang2026federated, dong2024fadngs}, its application to VAD remains relatively recent. CLAP~\cite{clap_cvpr2024} proposed an unsupervised VAD framework that uses clustering to generate pseudo-labels, though its reliance on local, heterogeneous data leads to inconsistent pseudo-labels that degrade global generalization. More recently, Fed-WSVAD~\cite{fedwsvad_aaai2025} introduced a multimodal prompt generation scheme to balance personalization with generalization. However, it does not explicitly enforce semantic consistency across the clients' feature spaces. These works highlight a core challenge in federated VAD, namely that models remain highly vulnerable to feature drift under significant data heterogeneity due to the lack of a shared semantic foundation.

\subsection{Visual–Language Adaptation and Prompt Learning in Federated Settings}\label{subsec:relatedwork_vlm}
Prompt learning has emerged as a parameter-efficient strategy for adapting large VLMs like CLIP~\cite{zsclip_icml2021} to downstream tasks. Methods like CoOp~\cite{zhou2022learning_coop} and CoCoOp~\cite{zhou2022conditional_cocop} learn continuous textual prompts, steering the model's predictions without fine-tuning the entire backbone. Subsequent approaches~\cite{wang2024vilt, bai2024prompt, wu2024vadclip} extend prompt learning to video or multi-modal domains, demonstrating improved transferability and efficiency. For instance, Temp-CLIP~\cite{temporalclip_eccv2022} refines temporal reasoning for video tasks by learning only a few prompt vectors and a lightweight temporal transformer \cite{vaswani2017attention}, while ActionCLIP~\cite{wang2023actionclip} adapts CLIP for video-based action recognition, demonstrating VLM adaptation for dynamic event understanding.

In federated learning, prompt-based adaptation is considered a highly practical approach due to its communication efficiency. Works such as FedCoOp~\cite{fedcoop_tmc2023}, pFedPrompt~\cite{guo2023pfedprompt}, and pFedPG~\cite{yang2023efficient} focus on aggregating or personalizing prompts across clients to handle data heterogeneity. However, a key limitation of these methods lies in their implicit assumption of a stable, universally shared label space. Because they are designed for standard classification tasks, they assume the semantics of a label are consistent across all clients. This assumption breaks down in VAD, where the very definition of an ``anomaly'' is context-dependent and varies significantly depending on a client's specific local environment.

In summary, while existing federated VAD and prompt learning methods successfully address privacy and communication efficiency, they collectively struggle with semantic misalignment under extreme non-IID conditions. To provide a comprehensive overview of this research landscape, we summarize the distinctions between our proposed FedVAR and state-of-the-art baselines in Table \ref{tab:related_work}. As highlighted in the table, existing frameworks (e.g., CLAP, Fed-WSVAD, FedCoOp) either lack a shared semantic foundation (Section \ref{subsec:relatedwork_vad}) or falsely assume a stable, homogeneous label space across edge devices (Section \ref{subsec:relatedwork_vlm}). To bridge this gap, our proposed FedVAR explicitly targets these limitations to enable fine-grained video anomaly recognition. Instead of naively averaging ambiguous prompts or misaligned model weights, FedVAR introduces a novel prototype-guided alignment mechanism (detailed in Section \ref{sec:method}). By anchoring the federated learning process to a globally shared representation of normality, our framework establishes the semantic consistency required for multi-class anomaly recognition in decentralized cyber-physical environments.

\section{Preliminaries: Centralized VAR}
\label{sec:preliminaries}
This section presents the essential background for our framework, specifically focusing on the AnomalyCLIP~\cite{anomalyclip} architecture, which serves as the foundational baseline for detection and recognition in the visual-textual embedding space.

\begin{figure*}[!t]
  \centering
  \begin{subfigure}[b]{0.3\textwidth}
    \centering
    \includegraphics[width=\textwidth]{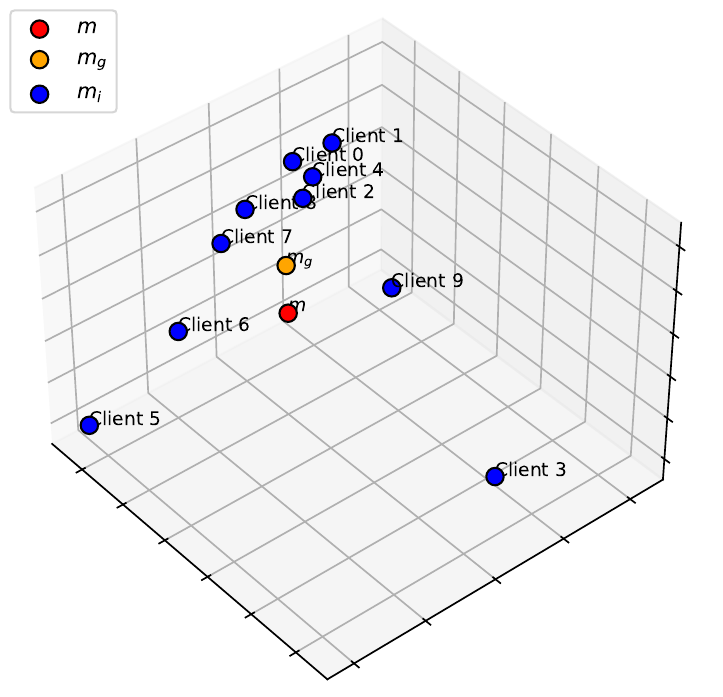}
    \caption{UCF-Crime}
    \label{fig:ucf_pca}
  \end{subfigure}
  \begin{subfigure}[b]{0.3\textwidth}
    \centering
    \includegraphics[width=\textwidth]{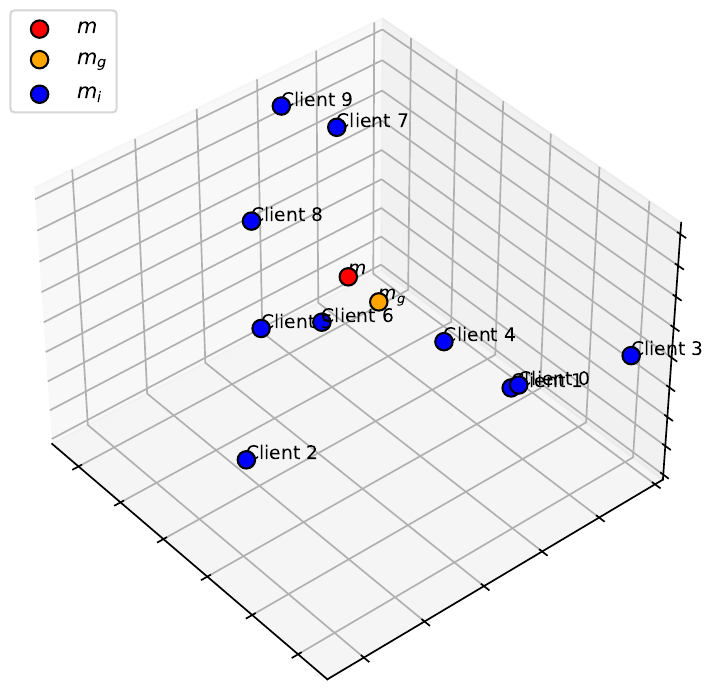}
    \caption{XD-Violence}
    \label{fig:xd_pca}
  \end{subfigure}
  \begin{subfigure}[b]{0.3\textwidth}
    \centering
    \includegraphics[width=\textwidth]{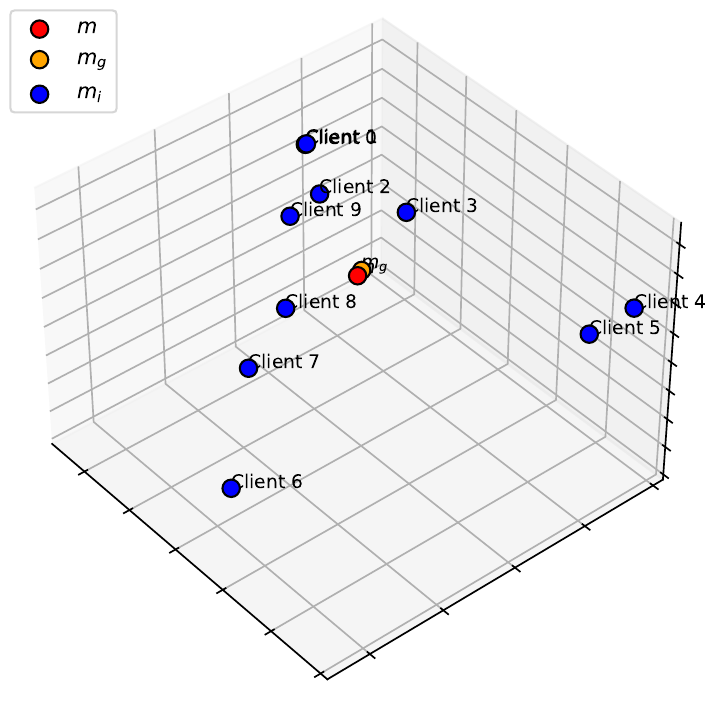}
    \caption{ShanghaiTech}
    \label{fig:sh_pca}
  \end{subfigure}
  \caption{3D visualization of normality prototypes, projected into three dimensions using Principal Component Analysis (PCA). Clients learn diverged local normality prototypes (\(m_{i}\)) on each dataset individually, leading to different anomaly direction vectors, while FedVAR learns a unified normality prototype (\(m_{g}\)) which is closest to the true normality prototypes (\(m\)) of the datasets.}
  \label{fig:pca_combined}
\end{figure*}

\subsection{Input Representation and Objective}
AnomalyCLIP learns to detect and recognize anomalies in videos by leveraging the shared visual-textual embedding space of Contrastive Language-Image Pre-training (CLIP)~\cite{zsclip_icml2021}. 
Let a video $V$ be represented as a sequence of frames $\{I_t\}_{t=1}^{T}$. Each frame $I_t$ is encoded by a frozen CLIP visual encoder $E_{\text{image}}(\cdot)$ into a spatial frame feature vector $x_t = E_{\text{image}}(I_t) \in \mathbb{R}^D$. 
The primary objective is to train a model parameterized by $\theta$ that, for any given frame feature $x_t$, can simultaneously predict: (i) a binary anomaly probability $p_A(x_t) \in [0,1]$ indicating the likelihood of an anomaly occurring at frame $I_t$ (Detection), and (ii) a conditional class probability $p_{c \mid A}(x_t)$ over a global vocabulary of anomaly classes $c \in \mathcal{C}$ (Recognition).
In weakly-supervised scenarios, these frame-level predictions are learned from video-level labels under the Multiple Instance Learning (MIL)~\cite{ucfcrime_sultani2018real} paradigm, which assumes that an anomalous video contains at least one anomalous frame.

\subsection{Centralized Normality Prototype and Feature Re-centering}
In weakly-supervised VAR, distinguishing anomalous events heavily relies on establishing a reliable baseline of normality. AnomalyCLIP achieves this by calculating a centralized ``normality prototype'' $m \in \mathbb{R}^D$, which represents the average representation of all normal frames across the centralized dataset:
\begin{equation}
m = \frac{1}{|N|} \sum_{I \in N} E_{\text{image}}(I),    
\end{equation}
where $N$ denotes the set of all normal frames and $|N|$ is the total count of these frames. 
Using this prototype, each frame feature $x_t$ is re-centered as $x'_t = x_t - m$. This transformation clusters normal frames around the origin, forcing anomalous frames to deviate from it. This separation enhances the model's capacity to discriminate between normal and abnormal patterns.

\subsection{Vision-Text Prompt Learning Module}
To identify the specific type of anomaly (e.g., robbery, traffic accident), a Vision-Text Prompting (VTP) module $\mathcal{S}$ associates re-centered frame features with textual labels in the CLIP embedding space. Instead of hand-crafted prompts, continuous prompt learning is adopted. A set of learnable context tokens $t^{ctx}$ is concatenated with discrete class-specific tokens $t^c$ representing each anomaly class $c \in \mathcal{C}$ (e.g., the word ``violence'').
These prompts are passed through a frozen CLIP text encoder $E_{\text{text}}(\cdot)$ and subsequently re-centered by subtracting the normality prototype $m$, yielding textual anomaly direction vectors $d_c$:
\begin{equation}
d_c = E_{\text{text}}\big([t^{ctx}, t^c]\big) - m.
\end{equation}
The projection of a re-centered frame feature $x'_t$ onto the direction vector $d_c$ measures the likelihood of frame $I_t$ belonging to class $c$:
\begin{equation}
\label{eq:bn_projection}
P(x'_t, d_c) = \mathrm{BN}\!\left(\frac{x'_t \cdot d_c}{\|d_c\|}\right),
\end{equation}
where $\mathrm{BN}$ denotes batch normalization. The VTP module output representing the raw likelihood distribution across all predefined categories is given by:
\begin{equation}
\label{eq:selector_output}
\mathcal{S}(x'_t) = [\, P(x'_t, d_1), \dots, P(x'_t, d_{|\mathcal{C}|}) \,] \in \mathbb{R}^{|\mathcal{C}|}.
\end{equation}

\subsection{Temporal Module and Joint Prediction}
Since anomalies naturally unfold over time, a Temporal module $\mathcal{T}(\{x'_t\}_{t=1}^{T})$ is introduced to process the sequence of re-centered frame features $\{x'_t\}_{t=1}^{T}$ and output an anomaly probability for each frame-feature $p_A(x'_t) \in [0,1]$.
To achieve a unified prediction, the framework integrates the predictions of the Temporal and VTP modules. The joint probability $p_{A,c}(x'_t)$ of the frame containing a specific anomaly class $c$ is derived as:
\begin{equation}
\label{eq:joint_prob_central}
p_{A,c}(x'_t) = p_A(x'_t) \times p_{c \mid A}(x'_t),
\end{equation}
where $p_{c \mid A}(x'_t) = \mathrm{softmax}(\mathcal{S}(x'_t))$. This formulation enables the baseline model to simultaneously determine whether an anomalous event is occurring and resolve its specific semantic type.

\section{Methodology}
\label{sec:method}
While centralized frameworks excel when all data is accessible, extending video anomaly recognition to decentralized, privacy-preserving environments poses significant statistical challenges. In this section, we present the federated formulation of the task, analyze the challenge of semantic misalignment, and detail our proposed FedVAR framework.

\begin{figure*}[!t]
    \centering
    \includegraphics[width=0.8\textwidth]{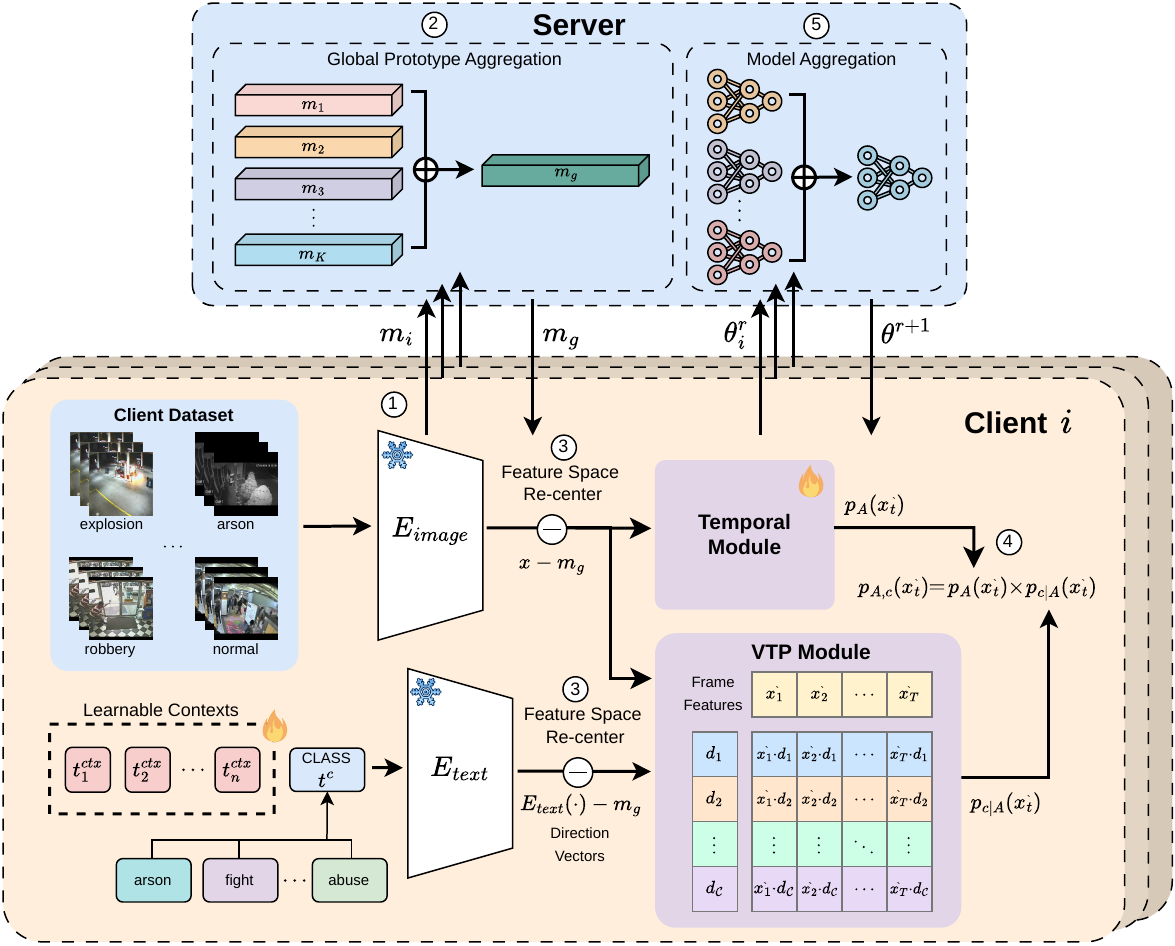}
    \caption{The architecture and workflow of our proposed FedVAR framework. Each client computes a local normality prototype (\(m_{i}\)) from its private normal data. These local prototypes are uploaded to the server and aggregated into a global prototype (\(m_{g}\)). The server distributes this shared anchor back to all clients. During local training, each client uses the global prototype \(m_{g}\) to re-center both visual and textual features, ensuring all clients operate in an aligned space. The trainable modules (VTP and Temporal) are updated locally. The updated model parameters (\(\theta_{i}^{r}\)) are sent to the server for aggregation, and the new global model (\(\theta^{r+1}\)) is distributed for the next round.}
    \label{fig:main_figure}
\end{figure*}

\subsection{Semantic Misalignment in Federated Video Anomaly Recognition}
\label{sec:fl_challenge_var}
We extend the weakly-supervised VAR problem to the FL setting, which involves a central server and a set of $K$ decentralized clients. Each client $i \in \{1, \dots, K\}$ holds a private local dataset $\mathcal{D}_i = \{(V_{ij}, c_{ij})\}_{j=1}^{|\mathcal{D}_i|}$, where $V_{ij}$ is the $j$-th video sample belonging to client $i$, weakly annotated with a video-level label $c_{ij} \in \mathcal{C} \cup \{\text{normal}\}$. Here, $\mathcal{C}$ denotes the global vocabulary of anomaly classes.

Due to the decentralized nature of data collection, these datasets are typically non-IID~\cite{lu2024federated}, exhibiting significant statistical heterogeneity across different clients. Clients capture distinct environmental contexts (e.g., indoor retail stores versus outdoor highway intersections), leading to highly divergent definitions of ``normal'' video streams. Furthermore, clients often observe disjoint subsets of the global anomaly classes $\mathcal{C}$. For instance, a traffic-monitoring client might exclusively observe ``accident'' anomalies, whereas a retail-monitoring client might only record ``robbery'' and ``shoplifting.''

Under this non-IID environment and weakly-supervised setting, directly applying the video anomaly recognition task is problematic. Figure~\ref{fig:pca_combined} illustrates that if each client $i$ independently computes its local normality prototype ($m_i$) from its distinct normal data distributions, these prototypes diverge substantially in the CLIP embedding space. We define this cross-client divergence as semantic misalignment, where each client develops a different interpretation of ``normal''. Consequently, if each client re-centers its visual and textual features based solely on its disjoint local $m_i$, the resulting anomaly direction vectors $d_c$ will point in entirely different directions for the same underlying anomaly class. As a result, the shared trainable parameters fail to generalize across the federation, leading to degraded cross-client anomaly recognition performance.
This challenge is further exacerbated by the rarity and diversity of anomalous events, which prevent any single client from learning a robust and universally applicable representation of normal and abnormal behavior.

\subsection{FedVAR: Cross-Modal Alignment and Joint Prediction}
\label{sec:fedvar_framework}
To address the semantic misalignment identified in Section~\ref{sec:fl_challenge_var}, FedVAR introduces a federated prototype alignment mechanism that constructs a shared global normality prototype $m_g \in \mathbb{R}^D$. This prototype serves as a common semantic anchor, ensuring that all clients operate within a consistent feature space despite heterogeneous local distributions. 
As illustrated in Figure~\ref{fig:main_figure}, FedVAR first performs cross-modal alignment through a shared global prototype, and then performs joint local training using re-centered visual and textual representations.

\paragraph{Cross-Modal Alignment via Global Prototype}
Prior to starting the iterative federated training rounds, each client computes its local normality prototype $m_i$ by averaging the visual features of its normal frames using the frozen CLIP image encoder $E_{\text{image}}(\cdot)$. Let $N_i$ denote the set of normal frames within the private dataset $\mathcal{D}_i$, and $|N_i|$ be the total count of these normal frames. The local prototype is formulated as:
\begin{equation}
m_i = \frac{1}{|N_i|} \sum_{I \in N_i} E_{\text{image}}(I).
\end{equation}
Clients transmit their local prototypes $m_i$ and the corresponding normal frame counts $|N_i|$ to the central server. The server then aggregates these private local representations into a shared global normality prototype $m_g$ using a weighted average:
\begin{equation}
\label{eq:global_norm_proto}
m_g = \frac{\sum_{i=1}^{K} |N_i| \cdot m_i}{\sum_{i=1}^{K} |N_i|}.
\end{equation}
By accounting for the sizes of the local normal subsets, $m_g$ establishes a unified, semantically balanced understanding of normality across the entire federation. Following the aggregation, the server broadcasts the global prototype $m_g$ back to all clients. From this point onward, every client $i$ transforms both its visual frame features and textual prompt embeddings using the shared global normality prototype $m_g$:
\begin{align}
x'_t &= E_{\text{image}}(I_t) - m_g, \\
d_c  &= E_{\text{text}}\big([t^{ctx}, t^c]\big) - m_g.
\end{align}
This uniform alignment ensures that all participating clients evaluate visual and textual representations relative to an identical and aligned semantic origin in the CLIP space. By anchoring all client representations to the same global semantic origin, the resulting anomaly direction vectors become consistently aligned and comparable across heterogeneous clients.

\paragraph{Joint Local Training through Re-centered Features}
After establishing the globally aligned feature space, FedVAR collaboratively optimizes only the trainable parameters $\theta=\{t^{ctx}, \mathcal{T}\}$, consisting of the learnable context tokens in the VTP module and the Temporal module parameters.
At communication round $r$, each participating client $i$ initializes its local model with the current global parameters $\theta^{r}$ and performs local optimization on its private dataset $\mathcal{D}_i$ for a fixed number of local epochs.

During each local update, the client first performs joint anomaly prediction using the globally aligned representations through the VTP module and the Temporal module. Specifically, the projection of the globally re-centered visual frame feature $x'_t$ onto the aligned anomaly direction vector $d_c$ yields the raw class-conditional likelihoods, from which the class-conditional probability distribution $p_{c \mid A}(x'_t)=\text{softmax}(\mathcal{S}(x'_t))$ is derived. Concurrently, the sequence of globally re-centered visual frame features $\{x'_t\}_{t=1}^{T}$ is processed by the Temporal module $\mathcal{T}$ to estimate the binary anomaly probability $p_A(x'_t)$.

By combining these two aligned signals, the final joint probability $p_{A,c}(x'_t)$ of the frame containing a specific anomaly class $c \in \mathcal{C}$ is obtained. 
Because the prediction is performed on the globally re-centered representations, the learnable context tokens $t^{ctx}$ and the Temporal module parameters $\mathcal{T}$ are optimized under a consistent semantic coordinate system across heterogeneous clients.

The resulting joint prediction is then used to compute the local loss function $\mathcal{L}_i(\theta; \mathcal{D}_i)$ on client $i$'s private dataset $\mathcal{D}_i$, where $\mathcal{L}_i$ measures the prediction error of the aligned anomaly detection and recognition tasks. By minimizing this local objective, client $i$ updates its trainable parameters $\theta_{i}^{r}$.

After local training, the updated parameters are transmitted to the central server. The server then aggregates the received client updates using standard federated averaging to produce the global model $\theta^{r+1}$ for the next communication round.
This iterative optimization, grounded in the globally aligned feature space, enables FedVAR to learn semantically consistent anomaly representations across highly heterogeneous clients.

\section{Convergence Analysis}
\label{sec:convergence_analysis}

To rigorously analyze the convergence behavior of the proposed FedVAR framework, we formalize the properties of the objective functions, the stochastic gradients, and the local training procedure. The joint trainable parameters are denoted as $\theta = \{t^{ctx}, \mathcal{T}\}$, encompassing the textual prompt tokens and the temporal module. Let $F_i(\theta; m) = \mathcal{L}_i(\theta; \mathcal{D}_i, m)$ denote the local objective function evaluated with a generic prototype shift variable $m$. The global objective optimized by our framework is given by $F(\theta; m_g) = \sum_{i=1}^K p_i F_i(\theta; m_g)$, where $m_g$ is the fixed global normality prototype and $p_i = \frac{|\mathcal{D}_i|}{\sum_{j=1}^K |\mathcal{D}_j|}$.

Consistent with the local-epoch training procedure described in Section~\ref{sec:fedvar_framework}, each communication round $r$ proceeds as follows: every client $i$ initializes $\theta_i^{r,0} = \theta^r$ and performs $E$ local stochastic gradient steps,
\begin{equation}
    \theta_i^{r,e+1} = \theta_i^{r,e} - \eta\, \nabla F_i(\theta_i^{r,e}; m_g, \xi_i^{r,e}), \quad e = 0, \dots, E-1,
\end{equation}
where $\eta$ is the local learning rate and $\xi_i^{r,e}$ is an independently sampled mini-batch. The server aggregates $\theta^{r+1} = \sum_{i=1}^K p_i \theta_i^{r,E}$. This is standard multi-step Federated Averaging, with $E$ denoting the total number of local gradient steps a client performs per communication round.

We make the following standard assumptions regarding the loss surface and gradient properties, commonly adopted in the federated optimization literature:

\textbf{Assumption 1} (\textit{Joint $L$-Smoothness}).
The local objective functions $F_i(\theta; m_g)$ are $L$-smooth with respect to the joint parameter space $\theta = \{t^{ctx}, \mathcal{T}\}$. That is, for all $i \in \{1, \dots, K\}$ and for any $\theta_1, \theta_2$, there exists a constant $L > 0$ such that:
\begin{equation}
    \|\nabla F_i(\theta_1; m_g) - \nabla F_i(\theta_2; m_g)\| \le L \|\theta_1 - \theta_2\|.
\end{equation}

\textbf{Assumption 2} (\textit{Bounded Intra-client Variance}).
Let $\xi \sim \mathcal{D}_i$ denote a uniformly sampled mini-batch from the local dataset of client $i$. The stochastic gradient $\nabla F_i(\theta; m_g, \xi)$ is an unbiased estimator of the true full-batch local gradient, with variance bounded by a constant $\sigma^2 > 0$:
\begin{align}
\mathbb{E}_\xi[\nabla F_i(\theta; m_g, \xi)] &= \nabla F_i(\theta; m_g),
\end{align}
\begin{align}
\mathbb{E}_\xi \left[\|\nabla F_i(\theta; m_g, \xi) - \nabla F_i(\theta; m_g)\|^2\right] &\le \sigma^2,
\end{align}
and mini-batch noise is independent across clients and across local steps within a client.

\textit{Remark:} To isolate the theoretical effects of our prototype alignment, we treat the Batch Normalization (BN) statistics within the VTP module as globally fixed constants. As demonstrated by prior work on FedBN~\cite{li2021fedbn}, locally computed BN statistics on non-IID data introduce a distinct source of feature shift. Our guarantees therefore characterize the aligned optimization landscape under this standard idealization, factoring out BN-induced drift to focus purely on the mitigation of semantic misalignment.

\textbf{Assumption 3} (\textit{Bounded Inter-client Gradient Divergence}).
To quantify the baseline statistical data heterogeneity across the federation, we bound the divergence of the local gradients evaluated at the globally aligned prototype $m_g$. There exists a constant $\Gamma_{\text{global}} \ge 0$ such that for any $\theta$:
\begin{equation}
    \sum_{i=1}^K p_i \|\nabla F_i(\theta; m_g) - \nabla F(\theta; m_g)\|^2 \le \Gamma_{\text{global}}^2.
\end{equation}

\textbf{Assumption 4} (\textit{$L_m$-Lipschitz Gradient w.r.t. Prototype Shift}).
The local gradients are $L_m$-Lipschitz continuous with respect to the prototype shift variable $m$. That is, for a fixed parameter $\theta$, and for any prototypes $m_1, m_2$, there exists a constant $L_m > 0$ such that:
\begin{equation}
    \|\nabla F_i(\theta; m_1) - \nabla F_i(\theta; m_2)\| \le L_m \|m_1 - m_2\|.
\end{equation}
\textit{Justification:} Writing $g_c(m) = \frac{(x-m)\cdot(e_c-m)}{\|e_c-m\|}$ for fixed $x = E_{\text{image}}(I_t)$ and $e_c = E_{\text{text}}([t^{ctx},t^c])$ — note both terms in the numerator depend on $m$ — $g_c$ is continuously differentiable on any domain where $\|e_c - m\| \ge \epsilon > 0$, with $\|\nabla_m g_c\|$ finite and bounded in terms of $\epsilon$ and the (bounded) norms of $x, e_c$. Since $m_g$ and every $m_i$ are convex combinations of CLIP normal-frame embeddings and are computed independently of the class prompt tokens $t^c$, we assume this separation holds throughout training.

\textbf{Lemma 1} (\textit{Prototype-Induced Gradient Divergence Bound}).
\textit{Let $\Gamma_{\text{global}}^2$ be the base gradient divergence evaluated at the globally aligned prototype $m_g$. Suppose clients instead naively utilize their unaligned local prototypes $m_i$, and let the corresponding unaligned local-prototype gradient divergence be defined as:}
\begin{equation}
    \Gamma_{\text{local}}^2 = \sum_{i=1}^K p_i \|\nabla F_i(\theta; m_i) - \nabla F_{\text{local}}(\theta)\|^2,
\end{equation}
\textit{where $\nabla F_{\text{local}}(\theta) = \sum_{i=1}^K p_i \nabla F_i(\theta; m_i)$. Under Assumptions 3 and 4, the unaligned gradient divergence is bounded by:}
\begin{equation}
    \Gamma_{\text{local}}^2 \leq 2\Gamma_{\text{global}}^2 + 2 L_m^2 \sum_{i=1}^K p_i \|m_i - m_g\|^2.
\end{equation}

\textbf{Proof.}
Since $\nabla F_{\text{local}}(\theta)$ is the exact weighted mean of the local gradients $\nabla F_i(\theta; m_i)$, replacing it with any arbitrary reference point — specifically the global aggregated gradient evaluated at $m_g$, denoted $\nabla F(\theta; m_g)$ — yields an upper bound via the fundamental property of variance:
\begin{equation}
    \Gamma_{\text{local}}^2 \leq \sum_{i=1}^K p_i \|\nabla F_i(\theta; m_i) - \nabla F(\theta; m_g)\|^2.
\end{equation}
Adding and subtracting $\nabla F_i(\theta; m_g)$ inside the norm and applying $\|A+B\|^2 \le 2\|A\|^2+2\|B\|^2$:
\begin{align}
    \Gamma_{\text{local}}^2 &\leq 2\sum_{i=1}^K p_i \|\nabla F_i(\theta; m_i) - \nabla F_i(\theta; m_g)\|^2 \nonumber \\ &\quad 
    + 2\sum_{i=1}^K p_i \|\nabla F_i(\theta; m_g) - \nabla F(\theta; m_g)\|^2.
\end{align}
By Assumption 3, the second term is bounded by $2\Gamma_{\text{global}}^2$, and by Assumption 4, the first term is bounded by $2L_m^2\sum_i p_i \|m_i-m_g\|^2$. $\hfill\square$

\textbf{Practical Implications for FedVAR:}
This lemma shows that a framework relying on unaligned local prototypes $m_i$ carries an additional, non-negative penalty term scaling with $\sum_i p_i\|m_i-m_g\|^2$ in its gradient-divergence bound. This does not by itself prove $\Gamma_\text{local}^2$ is larger in every instance, but it shows that unaligned re-centering cannot be guaranteed better than aligned re-centering, and its theoretical guarantee is strictly worse whenever the misalignment term dominates. Because FedVAR fixes $m_g$ prior to iterative optimization, it optimizes $F(\theta;m_g)$ directly, whose heterogeneity is governed solely by $\Gamma_\text{global}^2$, with Theorem~1 below reflects this by construction. Lemma~1's role is to make precise, via Proposition~1 and Corollary~1, how much worse the guarantee for an unaligned baseline would be.

\textbf{Lemma 2} (\textit{Bounded Client Drift}).
\textit{Suppose Assumptions 1--3 hold and $\eta \le \frac{1}{8LE}$. Then for every client $i$ and every local step $e \in \{0,\dots,E-1\}$ within round $r$,}
\begin{equation}
\begin{multlined}
\sum_{i=1}^K p_i\, \mathbb{E}\|\theta_i^{r,e} - \theta^r\|^2
\le \mathcal{O}\!\left(E\eta^2\sigma^2\right) \\
+ \mathcal{O}\!\left(
E^2\eta^2\left(
\Gamma_\text{global}^2
+ \mathbb{E}\|\nabla F(\theta^r;m_g)\|^2
\right)\right).
\end{multlined}
\end{equation}

\textit{Proof Sketch.} By definition, $\theta_i^{r,e}-\theta^r = -\eta\sum_{e'=0}^{e-1}\nabla F_i(\theta_i^{r,e'};m_g,\xi_i^{r,e'})$. Applying Jensen's inequality across the $e \le E$ terms, adding and subtracting $\nabla F_i(\theta^r;m_g)$, and using the identity $\sum_i p_i\|\nabla F_i(\theta^r;m_g)\|^2 \le \Gamma_\text{global}^2 + \|\nabla F(\theta^r;m_g)\|^2$ (which follows from Assumption 3 via the same variance decomposition used in Lemma~1) together with Assumption 2's variance bound, an inductive argument over $e = 0,\dots,E-1$ can be formed. Because the global prototype $m_g$ remains fixed throughout the communication round, this induction is mathematically identical to the standard client drift bounds established in federated non-convex optimization. We refer readers to the exact inductive derivations provided in \cite{wang2019adaptive, yang2021achieving}, which yield the stated bound.

\textbf{Theorem 1} (\textit{Convergence of FedVAR on Non-Convex Objectives}).
\textit{Suppose Assumptions 1 through 3 hold. Assume the global objective function $F(\theta;m_g)$ is non-convex and bounded below by $F^*$. If federated training is run for $R \ge E$ communication rounds with $E$ local steps per round and a fixed global prototype $m_g$, using a local learning rate $\eta = c/\sqrt{ER}$ for a sufficiently small constant $c \le \frac{1}{8L}$, then the algorithm converges to a stationary point. Specifically, the average expected squared gradient norm is bounded by:}
\begin{equation}
    \frac{1}{R}\sum_{r=1}^R \mathbb{E}\left[\|\nabla F(\theta^r;m_g)\|^2\right] \le \mathcal{O}\!\left(\frac{1}{\sqrt{ER}}\right) + \mathcal{O}\!\left(\frac{E\,\Gamma_\text{global}^2}{R}\right).
\end{equation}

\textit{Proof Sketch.}
We analyze one communication round as a single composite update $\theta^{r+1} = \theta^r - \eta E \Delta^r$, where $\Delta^r = \frac{1}{E}\sum_i p_i \sum_{e=0}^{E-1}\nabla F_i(\theta_i^{r,e};m_g,\xi_i^{r,e})$. By $L$-smoothness (Assumption 1) and unbiasedness (Assumption 2),
\begin{equation}
\begin{multlined}
\mathbb{E}[F(\theta^{r+1};m_g)]
\le F(\theta^r;m_g) \\
- \eta E\langle \nabla F(\theta^r;m_g), \mathbb{E}[\Delta^r]\rangle 
+ \frac{L(\eta E)^2}{2}\mathbb{E}\|\Delta^r\|^2.
\end{multlined}
\end{equation}
Since every client starts round $r$ at the shared point $\theta^r$, $\sum_i p_i \nabla F_i(\theta^r;m_g) = \nabla F(\theta^r;m_g)$ exactly at $e=0$; the deviation of $\mathbb{E}[\Delta^r]$ from $\nabla F(\theta^r;m_g)$ is therefore driven entirely by drift accumulated over the remaining local steps, bounded by Lemma~2. Applying Young's inequality to isolate $-\frac{\eta E}{2}\|\nabla F(\theta^r;m_g)\|^2$, substituting Lemma~2's drift bound and the noise bound $\mathbb{E}\|\Delta^r - \mathbb{E}\Delta^r\|^2 \le \sigma^2/E$ (Assumption 2, independence across clients), and absorbing the resulting $\mathcal{O}(\eta^3L^2E^3)\|\nabla F(\theta^r;m_g)\|^2$ term into the descent term via $\eta \le \frac{1}{8LE}$, we obtain
\begin{equation}
\begin{multlined}
\mathbb{E}[F(\theta^{r+1};m_g)]
\le \mathbb{E}[F(\theta^r;m_g)]
- \frac{\eta E}{4}\mathbb{E}\|\nabla F(\theta^r;m_g)\|^2 \\
+ \mathcal{O}(\eta^2 E L\sigma^2)
+ \mathcal{O}(\eta^3E^3L^2\Gamma_\text{global}^2).
\end{multlined}
\end{equation}
Telescoping this inequality over $R$ rounds, dividing by $\eta E R/4$, and setting $\eta = c/\sqrt{ER}$ yields the stated bound. Because optimizing with a statically aligned $m_g$ maps exactly to standard non-convex FedAvg, this virtual-sequence telescoping step directly mirrors the canonical proofs in \cite{wang2019adaptive, yang2021achieving}, thereby absorbing the stochastic-noise contribution into the leading $\mathcal{O}(1/\sqrt{ER})$ term and isolating the aligned heterogeneity contribution in the $\mathcal{O}(E\Gamma_\text{global}^2/R)$ term.

\textbf{Proposition 1} (\textit{Convergence of the Unaligned Baseline}).
\textit{Suppose Assumptions 1, 2, and 4 hold, and suppose additionally that an analogous heterogeneity bound holds pointwise at each client's own local prototype, i.e.\ $\sum_i p_i \|\nabla F_i(\theta;m_i) - \nabla F_\text{local}(\theta)\|^2 \le \Gamma_\text{local}^2$ for all $\theta$ (the quantity defined in Lemma~1). Then, under the same step-size schedule, an identical argument to the proof of Theorem~1, in which each client $i$ evaluates local gradients at its own prototype $m_i$ throughout local training rather than the shared $m_g$, yields}
\begin{equation}
    \frac1R\sum_{r=1}^R \mathbb{E}\left[\|\nabla F_\text{local}(\theta^r)\|^2\right] \le \mathcal{O}\!\left(\frac{1}{\sqrt{ER}}\right) + \mathcal{O}\!\left(\frac{E\,\Gamma_\text{local}^2}{R}\right).
\end{equation}

\textbf{Corollary 1} (\textit{Communication Efficiency and Semantic Misalignment}).
\textit{Let $\Delta_m = \sum_{i=1}^K p_i \|m_i - m_g\|^2$ quantify the degree of semantic misalignment across the federation. By Lemma~1, $\Gamma_\text{local}^2 \le 2\Gamma_\text{global}^2 + 2L_m^2\Delta_m$. Combined with Theorem~1 and Proposition~1, the unaligned baseline's guaranteed $\mathcal{O}(\cdot/R)$ error floor is never smaller than, and grows without bound relative to, FedVAR's as $\Delta_m$ increases. Consequently, whenever $2L_m^2\Delta_m$ is large enough to dominate $\Gamma_\text{global}^2$, reaching the same target accuracy $\varepsilon$ in the $\mathcal{O}(\cdot/R)$ term requires strictly more theoretical communication rounds for the unaligned baseline than for FedVAR, by a factor that grows with $\Delta_m$.}

\textit{Interpretation:} This result formalizes the benefit of prototype alignment: the two convergence guarantees are separated by a term controlled by the prototype-misalignment term identified in Lemma~1, isolating alignment as the source of FedVAR's improved theoretical communication efficiency, given that Assumptions 1--4 and the pointwise heterogeneity bound of Proposition~1 hold for both methods.

\section{Experimental Setup}\label{sec:exp_setup}
We conduct extensive experiments to evaluate the effectiveness of FedVAR. This section introduces the datasets and partitioning strategies, the competing baselines, the evaluation metrics, and the implementation details.

\begin{figure*}[!t]
    \centering
    \includegraphics[width=0.8\textwidth]{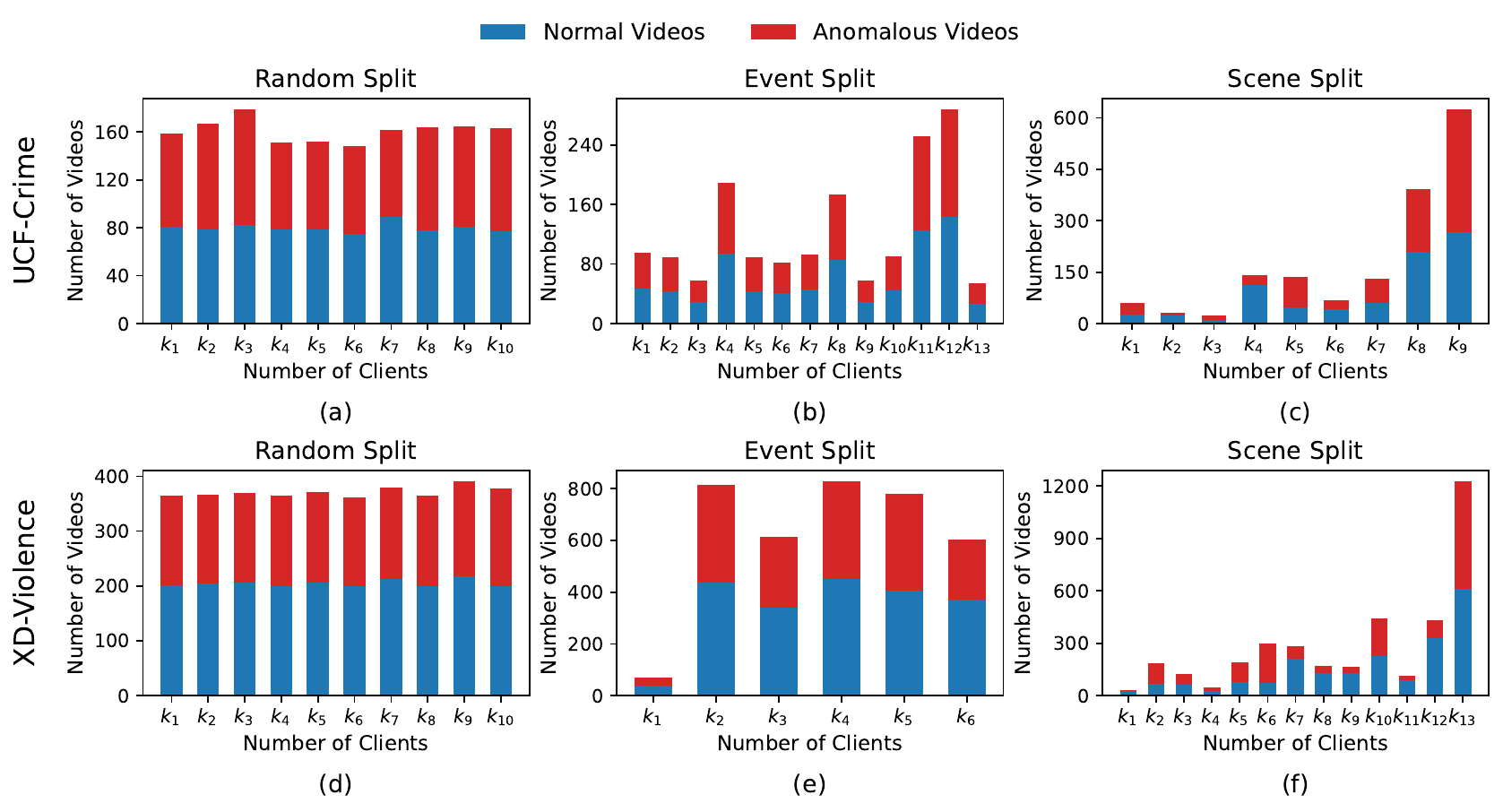}
    \caption{Client-wise data distribution under different partitioning strategies for UCF-Crime and XD-Violence. Each plot reports, for every client, the total number of videos with normal labels and anomalous labels counted separately, where the x-axis denotes client indices and the y-axis indicates the number of videos. Results are shown for three partitioning schemes: random split, event split, and scene split, illustrating varying degrees of statistical heterogeneity across clients. Note that these counts represent video-level labels, where anomalous events may occupy only a subset of frames in each video, rather than frame-level statistics.}
    \label{fig:client_data_distribution}
\end{figure*}

\subsection{Datasets and Partitioning Strategies}\label{subsec:dataset_partition}
We evaluate our approach on three widely-used video anomaly detection benchmarks. \textbf{UCF-Crime}~\cite{ucfcrime_sultani2018real} is a large-scale dataset comprising approximately 1,900 long untrimmed surveillance videos, covering 13 real-world anomaly categories (e.g., robbery, abuse, explosions) as well as normal activities. \textbf{XD-Violence}~\cite{xdviolence_wu2020not} is a larger and more diverse benchmark containing 4,754 untrimmed videos (217 hours in total), annotated with weak video-level labels and spanning multiple violent scenarios. \textbf{ShanghaiTech Campus}~\cite{shanghaitech_luo2017revisit} is a medium-scale dataset of 437 surveillance videos (307 normal, 130 anomalous) captured across 13 different scenes. Following~\cite{zhong2019graph}, we adopt its weakly supervised reorganization for evaluation.

To simulate diverse and realistic federated scenarios, we adopt three distinct data partitioning strategies, consistent with prior works~\cite{fedwsvad_aaai2025,clap_cvpr2024}. The client-wise sample distributions for each strategy are visualized in Figure~\ref{fig:client_data_distribution}.
\begin{enumerate}
    \item \textbf{Random Split:} Videos are sampled uniformly and distributed among clients. As shown in Figure~\ref{fig:client_data_distribution} (subplots (a) and (d)), this results in a relatively balanced setting where each client holds a similar number of normal and anomalous samples. However, some degree of heterogeneity may still exist in terms of the exact number of anomaly classes held by each client.
    \item \textbf{Event Split:} Videos are partitioned based on their anomaly class (Figure~\ref{fig:client_data_distribution}, subplots (b) and (e)), creating a highly heterogeneous non-IID environment where each client possesses data from only one anomaly type. This setup directly tests the model's ability to generalize across clients with non-overlapping class distributions. This setting represents an extreme but principled evaluation scenario for federated VAR, where semantic misalignment and class incompleteness are maximized.
    \item \textbf{Scene Split:} Videos are grouped by their recording location or scene. This represents the most realistic and challenging real-world scenario. As illustrated in Figure~\ref{fig:client_data_distribution} (subplots (c) and (f)), in this split, not only do clients receive a different number of videos, but the distribution of anomaly classes is also naturally skewed, as certain events are more likely to occur in specific scenes (e.g., road accidents on highways). This creates a complex non-IID setting that stresses both data and label heterogeneity, making fine-grained anomaly recognition substantially more difficult than binary detection.
\end{enumerate}

\subsection{Baselines}\label{subsec:baselines}
We benchmark our method against both centralized baselines (\textbf{ZS-CLIP} (ICML 2021)~\cite{zsclip_icml2021}, and \textbf{Temp-CLIP} (ECCV 2022)~\cite{temporalclip_eccv2022}) adapted for FL, and recent FL methods (\textbf{FedCoOp} (TMC 2023)~\cite{fedcoop_tmc2023}, \textbf{PPVU} (DSC 2023)~\cite{ppvu_dsc2023}, \textbf{CLAP} (CVPR 2024)~\cite{clap_cvpr2024}, and \textbf{Fed-WSVAD} (AAAI 2025)~\cite{fedwsvad_aaai2025}).
\textbf{ZS-CLIP}~\cite{zsclip_icml2021} serves as a zero-shot CLIP baseline using handcrafted prompts of the form ``this is a video of \{class\}.'' \textbf{Temp-CLIP}~\cite{temporalclip_eccv2022} extends CLIP with temporal modeling for video understanding. \textbf{FedCoOp}~\cite{fedcoop_tmc2023} adapts cooperative prompt learning (CoOp)~\cite{zhou2022learning_coop} to federated settings, enabling personalized text prompts for each client. \textbf{PPVU}~\cite{ppvu_dsc2023} is a transformer-based federated video understanding framework designed for heterogeneous client environments. \textbf{CLAP}~\cite{clap_cvpr2024} introduces pseudo-label generation and refinement for unsupervised federated VAD. Finally, \textbf{Fed-WSVAD}~\cite{fedwsvad_aaai2025} represents the current state-of-the-art in federated weakly supervised VAD, dynamically generating CLIP prompts from global textual and local visual contexts. Together, these methods span a diverse spectrum of paradigms, including zero-shot inference, temporal extension, prompt learning, and federated adaptations, ensuring a comprehensive comparison.
As none of the existing methods directly address the VAR task, we repurpose the most relevant VAD methods to construct suitable VAR baselines. Specifically:
(1) For ZS-CLIP, Temp-CLIP, FedCoOp, and Fed-WSVAD, we achieve anomaly recognition by applying a softmax operation over the cosine similarities between the input frame feature $\mathbf{x}$ and the anomaly direction vectors derived from class-specific CLIP text embeddings. These methods operate in a weakly supervised manner.
(2) For CLAP, we utilize video-level labels to generate initial pseudo-labels while employing CLIP-extracted features as input representations. We replace the original classification head with a multi-class prediction layer to recognize anomaly categories at the frame level, training it with a cross-entropy objective.
For fair comparison across all baselines, we implement federated versions of non-FL models (ZS-CLIP and Temp-CLIP) using the standard FedAvg~\cite{mcmahan2017communication} aggregation protocol. All clients train locally using only their own data, while the server maintains and updates a unified global model evaluated on a shared test set.

\subsection{Evaluation Metrics}
We conduct evaluations for both VAD and VAR. Following~\cite{ucfcrime_sultani2018real, xdviolence_wu2020not, clap_cvpr2024}, VAD performance is measured at the frame level. For UCF-Crime and ShanghaiTech, we report the Area Under the Receiver Operating Characteristic Curve (AUC-ROC, hereafter simply referred to as AUC), which reflects the model’s ability to distinguish normal from anomalous frames independent of thresholds. For XD-Violence, where anomalies are heavily imbalanced, we follow the established protocol~\cite{xdviolence_wu2020not} and report Average Precision (AP) computed from the precision–recall curve. For VAR, we extend these metrics to the multi-class setting by computing class-wise AUC or AP for each anomaly category and report their mean (mAUC or mAP). This dual evaluation captures both binary detection capability and fine-grained anomaly classification performance.

\begin{table*}[!t]
\centering
\setlength{\tabcolsep}{2.5pt}
\caption{VAR results (mAUC \%) on the UCF-Crime dataset under the Random split. The best-performing method for each category is highlighted in \textbf{bold}, while the second-best result is indicated by \underline{underline}.}
\label{tab:var_ucfcrime}
\resizebox{\textwidth}{!}{
\begin{tabular}{lcccccccccccccc|c}
\hline
\textbf{Method} & \textbf{Abuse} & \textbf{Arrest} & \textbf{Arson} & \textbf{Assault} & \textbf{Burglary} & \textbf{Explosion} & \textbf{Fighting} & \textbf{Normal} & \textbf{RoadAcc.} & \textbf{Robbery} & \textbf{Shooting} & \textbf{Shoplifting} & \textbf{Stealing} & \textbf{Vandalism} & \textbf{mAUC} \\
\hline
ZS-CLIP \cite{zsclip_icml2021} & 34.74 & 87.93 & 75.95 & \underline{90.09} & 80.13 & 90.17 & 80.14 & 68.08 & 90.85 & 57.88 & 31.03 & 61.07 & 56.58 & 29.02 & 66.58 \\
Temp-CLIP \cite{temporalclip_eccv2022} & 59.50 & 75.30 & 78.18 & 74.07 & 69.28 & 84.13 & 72.82 & 74.04 & 85.73 & 66.80 & 58.25 & 76.36 & 83.46 & 68.59 & 73.27 \\
FedCoOp \cite{fedcoop_tmc2023} & 78.00 & 77.17 & \underline{93.75} & 90.01 & 80.50 & 91.62 & 86.51 & 79.77 & \textbf{94.92} & 78.43 & 64.06 & 85.10 & 92.54 & 81.29 & 84.15 \\
CLAP \cite{clap_cvpr2024} & 68.99 & 62.14 & 81.72 & 78.61 & 75.98 & 87.84 & 72.64 & 76.09 & 91.03 & 67.75 & 66.60 & 82.95 & 91.14 & 65.13 & 76.35 \\
Fed-WSVAD \cite{fedwsvad_aaai2025} & \textbf{88.00} & \underline{89.74} & 93.10 & 83.17 & \underline{85.51} & \textbf{95.44} & \underline{87.65} & \underline{83.04} & \underline{94.87} & \underline{83.17} & \underline{76.69} & \underline{89.95} & \textbf{97.50} & \underline{85.13} & \underline{88.46} \\
FedVAR (ours) & \underline{79.16} & \textbf{94.53} & \textbf{95.88} & \textbf{94.13} & \textbf{89.60} & \underline{93.54} & \textbf{89.18} & \textbf{85.95} & 93.58 & \textbf{88.58} & \textbf{86.90} & \textbf{90.57} & \underline{97.20} & \textbf{90.90} & \textbf{91.06} \\
\hline
\end{tabular}
}
\end{table*}

\begin{table*}[!t]
\centering
\setlength{\tabcolsep}{2.5pt}
\caption{VAR results (mAUC \%) on the ShanghaiTech dataset under the Random split. The best-performing method for each category is highlighted in \textbf{bold}, while the second-best result is indicated by \underline{underline}.}
\label{tab:var_shanghaitech}
\resizebox{\textwidth}{!}{
\begin{tabular}{lcccccccccccccccc|c}
\hline
\textbf{Method} & \textbf{Car} & \textbf{Chasing} & \textbf{Circuit} & \textbf{Fall} & \textbf{Fighting} & \textbf{Jumping} & \textbf{Monocycle} & \textbf{Normal} & \textbf{Push} & \textbf{Robbery} & \textbf{Running} & \textbf{Skateboard} & \textbf{Stoop} & \textbf{ThrowObj.} & \textbf{Vaudeville} & \textbf{Vehicle} & \textbf{mAUC} \\
\hline
ZS-CLIP \cite{zsclip_icml2021} 
& 63.25 & 49.05 & 11.51 & 44.66 & 85.61 & 22.58 & \textbf{92.81} & 56.90 & 42.57 & 38.59 & 54.22 & \underline{77.33} & 5.45 & 44.32 & 33.07 & 37.09 & 46.81 \\
Temp-CLIP \cite{temporalclip_eccv2022} 
& 74.33 & \underline{92.35} & 91.94 & 85.31 & 91.59 & 89.07 & 82.00 & \underline{92.35} & 87.10 & \underline{81.76} & \underline{71.94} & 74.98 & \underline{80.87} & \underline{86.73} & \textbf{95.61} & \textbf{86.93} & \underline{84.83} \\
FedCoOp \cite{fedcoop_tmc2023} 
& 52.44 & 86.56 & 87.49 & \underline{87.11} & \textbf{95.60} & \textbf{96.10} & 69.53 & 82.56 & \underline{92.35} & 64.07 & 51.69 & 56.84 & 75.53 & 81.49 & 88.72 & 81.47 & 77.80 \\
CLAP \cite{clap_cvpr2024} 
& 58.96 & 66.30 & 38.58 & 71.26 & 71.95 & 62.38 & 23.71 & 45.05 & 61.60 & 60.06 & 36.09 & 38.88 & 45.68 & 54.23 & 61.91 & 41.65 & 52.88 \\
Fed-WSVAD \cite{fedwsvad_aaai2025} 
& \underline{85.76} & \textbf{95.78} & \textbf{98.98} & 71.92 & \underline{93.49} & 88.41 & \underline{86.99} & 90.58 & 85.16 & 58.79 & 52.06 & 61.65 & 39.57 & \underline{83.42} & 90.25 & 83.68 & 78.39 \\
FedVAR (ours) 
& \textbf{91.70} & 88.11 & \underline{97.47} & \textbf{93.82} & 90.56 & \underline{95.90} & 84.07 & \textbf{94.81} & \textbf{96.04} & \textbf{91.95} & \textbf{90.64} & \textbf{92.78} & \textbf{95.30} & \textbf{92.82} & \underline{93.26} & \underline{86.20} & \textbf{92.04} \\
\hline
\end{tabular}
}
\end{table*}

\begin{table*}[!t]
\centering
\caption{VAR results (mAP \%) on the XD-Violence dataset under the Random split. The best-performing method for each category is highlighted in \textbf{bold}, while the second-best result is indicated by \underline{underline}.}
\label{tab:var_xdviolence}
\resizebox{0.65\textwidth}{!}{
\begin{tabular}{lccccccc|c}
\hline
\textbf{Method} & \textbf{Abuse} & \textbf{Car Accident} & \textbf{Explosion} & \textbf{Fighting} & \textbf{Normal} & \textbf{Riot} & \textbf{Shooting} & \textbf{mAP} \\
\hline
ZS-CLIP \cite{zsclip_icml2021} & 0.48 & \textbf{31.32} & 68.90 & 43.31 & 93.20 & 64.29 & 2.21 & 35.09 \\
Temp-CLIP \cite{temporalclip_eccv2022} & 0.90 & 20.43 & 47.73 & 34.23 & 95.14 & 81.93 & 5.37 & 31.77 \\
FedCoOp \cite{fedcoop_tmc2023} & 3.26 & 27.34 & \underline{69.06} & 59.42 & 96.77 & 91.40 & 14.69 & 44.20 \\
CLAP \cite{clap_cvpr2024} & 1.38 & 24.27 & 48.04 & 55.47 & 96.77 & 86.85 & 14.82 & 38.47 \\
Fed-WSVAD \cite{fedwsvad_aaai2025} & \textbf{13.87} & 28.73 & \textbf{70.16} & \underline{68.52} & \textbf{98.11} & \underline{92.66} & \underline{22.04} & \textbf{49.33} \\
FedVAR (ours) & \underline{5.00} & \underline{30.29} & 66.72 & \textbf{69.00} & \underline{97.74} & \textbf{93.10} & \textbf{23.46} & \underline{47.93} \\
\hline
\end{tabular}
}
\end{table*}

\subsection{Implementation Details}
We implement FedVAR using PyTorch \cite{ansel2024pytorch} with the Flower \cite{KUNDROO2025102276} framework for federated orchestration. The CLIP visual encoder (ViT-B/16) and text encoder are pre-trained and kept frozen during training, while learnable prompts and temporal module are optimized. In order to capture both short and long-range sequential dynamics in each video, we choose the Axial Transformer~\cite{ho2019_axial_transformer} as the temporal module. Frame features are extracted with dimension $D=512$. Each video is divided into $S=32$ segments with $F=16$ frames per segment. Mini-batches of size $B=32$ are balanced between normal and anomalous samples following the MIL formulation. We employ the AdamW \cite{loshchilovdecoupled} optimizer with learning rates ($\eta$) of $5\times 10^{-4}$, $1\times 10^{-2}$, and $1\times 10^{-2}$ for ShanghaiTech, UCF-Crime, and XD-Violence, respectively, and fix the prompt length to $t^{ctx} = 8$. We configure our federated experiments to align with established benchmarks~\cite{fedwsvad_aaai2025,clap_cvpr2024}. The number of clients (\( K \)) is determined by the partitioning strategy. For the Random split, we set \( K=10 \). For the Event and Scene splits, the number of clients is naturally defined by the number of unique anomaly classes or scenes in each dataset, respectively (e.g., 13 clients for UCF-Crime's event split). In our main experiments, we adopt a full participation setup by default, where all clients participate in training during every communication round, unless explicitly varied (e.g., in Section~\ref{subsec:ablation_cpr}). Each client performs $10$ local epochs with a batch size of 32, and the global model is trained for a total of 20 FL communication rounds. All reported results are the average of three independent runs with different random seeds.

\section{Results and Analysis}
In this section, we present extensive experiments to evaluate the performance of FedVAR against state-of-the-art baselines. We analyze results across datasets, data partitioning strategies, domain shift scenarios, unseen classes, and different FL configurations. Both quantitative and qualitative analyses are provided.

\subsection{Evaluation Against Baselines}\label{subsec:var_results}
We begin by evaluating the VAR performance of FedVAR against representative baselines. To establish a fair and direct comparison of the fundamental recognition capabilities of each model, the results presented in Tables~\ref{tab:var_ucfcrime}, \ref{tab:var_shanghaitech}, and \ref{tab:var_xdviolence} are all obtained under the Random split data partitioning scheme, as defined in Section~\ref{subsec:dataset_partition}. As detailed in Section~\ref{subsec:baselines}, all models (with the exception of PPVU~\cite{ppvu_dsc2023}) utilize the same frozen CLIP ViT-B/16 backbone to ensure that performance differences are attributable to the federated learning approach rather than the underlying feature extractor.

As reported in Table~\ref{tab:var_ucfcrime} for UCF-Crime, FedVAR achieves the highest mean AUC ($91.06$), outperforming FedCoOp by $+6.91\%$ and ZS-CLIP by $+24.48\%$. Notably, it delivers substantial gains in challenging categories such as \textit{Arrest ($94.53$)} and \textit{Shooting} ($86.90$). These categories represent challenging cases because anomalous cues are often brief and visually overlap with normal activities. This overlap complicates recognition in federated settings, where individual clients may observe only partial or biased views of these events, causing local normality prototypes to drift and leading to inconsistent anomaly direction vectors across clients. While existing methods such as FedCoOp or Fed-WSVAD attempt to aggregate model parameters or prompts, they do not correct this drift at the feature representation level. By enforcing a shared global normality prototype, FedVAR directly anchors all clients to a common semantic reference, reducing this cross-client semantic drift and allowing anomaly direction vectors to remain aligned even when anomaly classes are sparsely or unevenly distributed across clients. This explains why FedVAR yields larger relative improvements compared to methods such as FedCoOp or Fed-WSVAD, which rely solely on parameter aggregation without correcting the underlying feature space misalignment.

Table~\ref{tab:var_shanghaitech} summarizes the ShanghaiTech results, where FedVAR again establishes a clear advantage, achieving the best mean AUC ($92.04$). This corresponds to improvements of $+14.24\%$ over FedCoOp and $+45.23\%$ over ZS-CLIP. The dataset contains multiple anomaly categories captured from fixed surveillance viewpoints, where normal behavior is strongly dependent on the specific scene context. Under such conditions, learning a unified notion of normality across distributed training data becomes particularly difficult without an explicit alignment mechanism. The improved performance of FedVAR suggests that explicitly aligning representations related to normal behavior across clients contributes to more consistent recognition across anomaly categories.

In Table~\ref{tab:var_xdviolence} on XD-Violence, FedVAR achieves a mean AP of $47.93$. This represents an improvement of $+3.73\%$ over FedCoOp and $+12.84\%$ over ZS-CLIP, while remaining competitive with Fed-WSVAD. XD-Violence consists of exceptionally long, untrimmed videos in which anomalous events occupy only a small fraction of each video. In this extremely sparse setting, methods like Fed-WSVAD benefit from dynamic temporal mechanisms explicitly optimized for binary anomaly localization. While this grants Fed-WSVAD a marginal performance edge on this specific benchmark (as further discussed in Section~\ref{sec:discussion}), FedVAR nonetheless delivers strong recognition capabilities. This indicates that our proposed global prototype alignment strategy successfully preserves cross-client semantic consistency and remains highly effective even in challenging environments characterized by long video durations and sparse anomalous segments.

Overall, these results validate the effectiveness of our proposed framework. By combining temporal modeling with federated normality prototype alignment and prompt learning, FedVAR delivers consistent improvements over centralized CLIP-based approaches and federated baselines, establishing new state-of-the-art performance on UCF-Crime and ShanghaiTech, while remaining highly competitive on XD-Violence.

\begin{table}[!t]
\centering
\caption{VAD results for state-of-the-art methods and FedVAR under three data partitioning strategies: Random, Event, and Scene-based splits (AUC (\%) on UCF-Crime and AP (\%) on XD-Violence).}
\label{tab:vad_main_table}
\resizebox{1.0\columnwidth}{!}{
\begin{tabular}{l|l|cc|cc|cc}
\toprule
\textbf{Method} & \textbf{Feature} & \multicolumn{2}{c|}{\textbf{Random}} & \multicolumn{2}{c|}{\textbf{Event}} & \multicolumn{2}{c}{\textbf{Scene}} \\
 & & \textbf{UCF} & \textbf{XD} & \textbf{UCF} & \textbf{XD} & \textbf{UCF} & \textbf{XD} \\
\midrule
ZS-CLIP \cite{zsclip_icml2021} & CLIP & 68.08 & 51.93 & 68.08 & 51.93 & 68.08 & 51.93 \\
Temp-CLIP \cite{temporalclip_eccv2022} & CLIP & 74.04 & 62.81 & 73.09 & 60.88 & 72.73 & 64.47 \\
FedCoOp \cite{fedcoop_tmc2023} & CLIP & 79.77 & 64.14 & 72.53 & 40.98 & 78.09 & 75.44 \\
PPVU \cite{ppvu_dsc2023} & TimeSformer & 82.90 & - & - & - & - & - \\
CLAP \cite{clap_cvpr2024} & CLIP & 76.09 & 69.87 & 68.41 & 36.43 & 60.29 & 47.60 \\
Fed-WSVAD \cite{fedwsvad_aaai2025} & CLIP & 83.04 & \textbf{77.33} & 82.95 & 78.93 & 81.90 & \textbf{78.54} \\
\midrule
FedVAR (ours) & CLIP & \textbf{86.36} & 75.20 & \textbf{84.97} & \textbf{82.82} & \textbf{86.35} & 75.14 \\
\bottomrule
\end{tabular}
}
\end{table}

\subsection{Comparisons on Different Data Splits}
In order to measure how well each federated method maintains performance as the client data becomes increasingly non-IID, we now analyze the model's robustness to varying degrees of data heterogeneity. In this subsection, we evaluate the VAD performance across all three data partitioning strategies: Random, Event, and Scene splits, discussed in Section~\ref{subsec:dataset_partition}. The results are presented in Table~\ref{tab:vad_main_table}.

Under the Random split, FedVAR achieves an AUC of $86.36$ on UCF-Crime and an AP of $75.20$ on XD-Violence, surpassing FedCoOp by $+6.59\%$ (UCF) and $+11.06\%$ (XD), and outperforming Fed-WSVAD by $+3.32\%$ on UCF while remaining competitive on XD. Compared to ZS-CLIP, our approach delivers large improvements of $+18.28\%$ (UCF) and $+23.27\%$ (XD), confirming that simple handcrafted prompts fail to generalize in federated settings.

On the Event split, which introduces stronger distributional skew, FedVAR achieves $84.97$ (UCF) and $82.82$ (XD). This corresponds to gains of $+12.44\%$ (UCF) and $+41.84\%$ (XD) over FedCoOp, highlighting the ability of our method to adapt across clients with disjoint event categories. In this setting, baselines often suffer from severe client drift because local models overfit to their specific anomaly types, pulling the global model in conflicting semantic directions. By centering all features around a globally shared normality prototype prior to temporal modeling, FedVAR ensures that each client learns consistent anomaly direction vectors. This architectural choice prevents conflicting updates and explains our stronger cross-event generalization, particularly in the untrimmed videos of XD-Violence where we outperform Fed-WSVAD by $+3.89\%$.

For the most challenging Scene split, where anomalies are tied to particular backgrounds or environments, FedVAR achieves $86.35$ (UCF) and $75.14$ (XD). This not only outperforms FedCoOp by $+8.26\%$ (UCF) but also improves substantially over CLAP by $+26.06\%$ (UCF) and $+27.54\%$ (XD). Compared to Fed-WSVAD, our method yields higher AUC on UCF ($+4.45\%$) but slightly lower AP on XD ($-3.40\%$). The severe performance degradation observed in baseline methods under the Scene split occurs because ``normal'' background features heavily dominate the representations, causing misalignments across clients deployed in different physical locations. FedVAR’s prototype alignment naturally subtracts this scene-specific bias by re-centering the features, effectively isolating the dynamic anomalous events from the static background. 

Overall, these results demonstrate that FedVAR consistently maintains strong performance across diverse data splits. The robustness of our approach highlights that aggregating normality prototypes is not just a regularization technique, but a fundamental mechanism to correct the underlying feature-space divergence, strengthening generalization across highly non-IID events and scenes.

\begin{table}[!t]
\centering
\caption{Cross-domain generalization results when XD-Violence is used as the source dataset (AUC (\%) on UCF-Crime and ShanghaiTech, and AP (\%) on XD-Violence).}
\label{tab:unseen_data_xdviolence}
\resizebox{0.70\columnwidth}{!}{
\begin{tabular}{l|c|cc}
\toprule
\textbf{Method} & \multicolumn{1}{c|}{\textbf{Source}} & \multicolumn{2}{c}{\textbf{Target}} \\
& \textbf{XD} & \textbf{UCF} & \textbf{SHTech} \\
\midrule
ZS-CLIP \cite{zsclip_icml2021} & 51.93 & 68.08 & 56.90 \\
Temp-CLIP \cite{temporalclip_eccv2022} & 62.81 & 69.95 & 49.48 \\
FedCoOp \cite{fedcoop_tmc2023} & 64.14 & 72.27 & 61.09 \\
Fed-WSVAD \cite{fedwsvad_aaai2025} & \textbf{77.33} & 78.70 & 38.05 \\
\midrule
FedVAR (ours) & 75.20 & \textbf{81.66} & \textbf{48.23} \\
\bottomrule
\end{tabular}
}
\end{table}

\subsection{Cross-Domain Generalization}
We next examine how well FedVAR generalizes to unseen datasets when trained on a single source domain. This scenario evaluates robustness under cross-domain generalization settings, a critical challenge for practical deployment where anomaly distributions vary across environments. We adopt the random split and train a single shared prompt vector, then test transfer learning performance across UCF-Crime, XD-Violence, and ShanghaiTech. The results are reported in Tables~\ref{tab:unseen_data_xdviolence} and \ref{tab:unseen_data_ucfcrime}.

When trained on XD-Violence, FedVAR achieves 81.66 AUC on UCF-Crime and 48.23 AUC on ShanghaiTech, outperforming Fed-WSVAD by +2.96\% and 10.18\% respectively. Notably, while Fed-WSVAD achieves higher AP on the source dataset, its transfer to ShanghaiTech drops severely (38.05), whereas FedVAR remains more stable (48.23, +10.18\%). 
This sharp drop in baseline performance occurs because standard models tend to memorize the absolute visual appearance and background contexts of the source domain anomalies. In contrast, FedVAR learns anomalies as relative deviations from the shared normality prototype in the aligned feature space. Because ``normal'' behavioral patterns (e.g., pedestrians walking) are structurally more consistent across domains than anomalies (e.g., specific fighting styles), realigning client knowledge around this common notion of normality provides a powerful domain-invariant regularization.

Conversely, when trained on UCF-Crime, FedVAR reaches 60.20 AP on XD-Violence and 65.71 AUC on ShanghaiTech. While Fed-WSVAD achieves higher transfer performance on XD-Violence (likely due to its temporal mechanisms being specifically optimized for sparse anomaly localization in long untrimmed videos), FedVAR demonstrates stronger generalization on ShanghaiTech, where scene diversity and cross-domain semantic consistency become more critical. Compared to earlier CLIP-based baselines, our method provides substantial gains: +10.80\% over FedCoOp and +11.73\% over Temp-CLIP on ShanghaiTech. These results confirm that prototype aggregation not only enhances in-domain performance but also equips the global model with stronger transferability across disjoint datasets with drastically different visual conditions.

Overall, FedVAR demonstrates consistent robustness under domain shift, achieving the best or second-best performance in nearly all cases. Importantly, unlike Fed-WSVAD, which often sacrifices cross-domain stability to maximize source-domain localization, our approach balances in-domain performance with out-of-domain generalization. This proves that anchoring learning to a transferable representation of normality prevents models from overfitting to source-specific biases, allowing the framework to successfully generalize to entirely novel domains.

\begin{table}[!t]
\centering
\caption{Cross-domain generalization results when UCF-Crime is used as the source dataset (AUC (\%) on UCF-Crime and ShanghaiTech, and AP (\%) on XD-Violence).}
\label{tab:unseen_data_ucfcrime}
\resizebox{0.70\columnwidth}{!}{
\begin{tabular}{l|c|cc}
\toprule
\textbf{Method} & \multicolumn{1}{c|}{\textbf{Source}} & \multicolumn{2}{c}{\textbf{Target}} \\
& \textbf{UCF} & \textbf{XD} & \textbf{SHTech} \\
\midrule
ZS-CLIP \cite{zsclip_icml2021} & 68.08 & 51.93 & 56.90 \\
Temp-CLIP \cite{temporalclip_eccv2022} & 74.04 & 48.78 & 53.98 \\
FedCoOp \cite{fedcoop_tmc2023} & 79.77 & 58.22 & 54.91 \\
Fed-WSVAD \cite{fedwsvad_aaai2025} & 83.04 & \textbf{65.42} & 59.46 \\
\midrule
FedVAR (ours) & \textbf{86.36} & 60.20 & \textbf{65.71} \\
\bottomrule
\end{tabular}
}
\end{table}

\begin{table}[!t]
\centering
\caption{Unseen anomaly class generalization results (AUC (\%) on UCF-Crime and ShanghaiTech, and AP (\%) on XD-Violence).}
\label{tab:unseen_classes}
\resizebox{0.70\columnwidth}{!}{
\begin{tabular}{l|c|c|cc}
\toprule
\textbf{Method} & \textbf{UCF} & \textbf{XD} & \textbf{SHTech} \\
\midrule
ZS-CLIP \cite{zsclip_icml2021} & 68.08 & 51.93 & 56.90 \\
Temp-CLIP \cite{temporalclip_eccv2022} & 76.28 & 65.78 & 92.51 \\
FedCoOp \cite{fedcoop_tmc2023} & 76.26 & 60.19 & 82.45 \\
Fed-WSVAD \cite{fedwsvad_aaai2025} & 80.72 & \textbf{77.69} & 89.35 \\
\midrule
FedVAR (ours) & \textbf{83.06} & 73.08 & \textbf{93.69} \\
\bottomrule
\end{tabular}
}
\end{table}

\begin{table*}[!t]
\centering
\caption{Ablation study on the effect of using global and local normality prototypes on UCF-Crime and XD-Violence (AUC (\%) on UCF-Crime and AP (\%) on XD-Violence).}
\label{tab:ablation_prototype}
\resizebox{0.70\textwidth}{!}{
\begin{tabular}{cc|cccc|cccc}
\toprule
\multicolumn{2}{c|}{Normality Prototype} & \multicolumn{4}{c|}{UCF} & \multicolumn{4}{c}{XD} \\ 
\cmidrule(lr){1-2} \cmidrule(lr){3-6} \cmidrule(lr){7-10}
\textbf{\(m_{g}\)} & \textbf{\(m_{i}\)} &
\textbf{Random} & \textbf{Event} & \textbf{Scene} & \textbf{AVG} &
\textbf{Random} & \textbf{Event} & \textbf{Scene} & \textbf{AVG} \\ 
\midrule
 & \checkmark & 85.95 & 84.13 & 83.82 & 84.63 & 69.45 & 73.83 & 72.59 & 71.95 \\
\checkmark &  & \textbf{86.36} & \textbf{84.97} & \textbf{86.35} & \textbf{85.89} & \textbf{75.20} & \textbf{82.82} & \textbf{75.14} & \textbf{77.72} \\
\bottomrule
\end{tabular}
}
\end{table*}

\subsection{Generalization to Unseen Classes}
We further evaluate the generalization ability of FedVAR to anomaly categories that are entirely absent during training, a crucial capability for real-world deployment. For this experiment, we partitioned the anomaly classes of each dataset into two disjoint sets of approximately equal size: ``base'' (seen during training) and ``new'' (unseen, used for testing). This one-time split was performed randomly and then held fixed for all evaluated methods to ensure a consistent and fair comparison. Videos from base classes were used for federated training under the Event split, where each client owns a completely disjoint anomaly class, simulating a challenging scenario where the model must learn a generalizable concept of ``abnormality'' without seeing all event types. A single shared prompt vector was optimized during training. The results, reported in Table~\ref{tab:unseen_classes}, measure the model's ability to extrapolate to the ``new'' classes.

On UCF-Crime, FedVAR achieves an AUC of 83.06, outperforming Fed-WSVAD by +2.34\% and surpassing Temp-CLIP by +6.78\%. On ShanghaiTech, our approach achieves 93.69 AUC, marking a +1.18\% gain over Temp-CLIP and a +4.34\% improvement over Fed-WSVAD. On XD-Violence, while Fed-WSVAD attains the highest AP (77.69), FedVAR remains competitive at 73.08, exceeding Temp-CLIP and FedCoOp by +7.30\% and +12.89\%, respectively.

These results highlight two key observations. First, compared to prior CLIP-based federated methods such as FedCoOp, our approach consistently demonstrates strong improvements due to our joint vision-text temporal formulation. As FedVAR aligns the visual features around the global normality prototype, the textual prompts primarily learn the geometric property of ``deviation'' rather than memorizing the exact visual signatures of the training classes. Consequently, when an unseen anomaly occurs, it naturally projects strongly along the learned deviation vectors in the shared CLIP space, allowing the model to better generalize to unseen anomalies.
Second, while Fed-WSVAD achieves stronger performance on XD-Violence, potentially benefiting from its temporal modeling mechanisms for localizing short anomalous events within long untrimmed videos, FedVAR demonstrates stronger generalization on UCF-Crime and ShanghaiTech, which appear to benefit more from robust semantic representations and cross-client feature alignment under heterogeneous federated settings.

Overall, this confirms that our federated prototype aggregation strategy successfully promotes extrapolation beyond known anomaly classes. By learning a generalizable representation of ``abnormality'' as a deviation from an aligned normal representation, FedVAR provides a robust solution for real-world systems where novel anomalous events continually emerge.

\subsection{Ablation Studies}
\subsubsection{Ablation Study on Normality Prototype}
To validate the effectiveness of the proposed global normality prototype aggregation, we perform ablation experiments by removing the federated prototype alignment module and training clients using only their local normality prototypes $m_i$. Table~\ref{tab:ablation_prototype} reports the results across UCF-Crime and XD-Violence under the three partitioning schemes. When using only local normality prototypes $m_i$, performance consistently drops across all settings, indicating that without the shared anchor provided by the global prototype $m_g$, client models learn semantically misaligned feature spaces, leading to a degraded global model. In contrast, introducing the global normality prototype $m_g$ yields clear and consistent gains, with an average improvement of +1.26\% AUC on UCF-Crime and +5.77\% on XD-Violence. The larger improvement on XD-Violence, which exhibits greater domain diversity, highlights the global prototype’s role in stabilizing feature alignment under highly heterogeneous client distributions.

These findings demonstrate that our proposed prototype-based alignment mechanism is the key driver of FedVAR's robustness. Naively training with local prototypes is insufficient to overcome the semantic misalignment inherent in federated VAR. By establishing a shared, global representation of normality, our framework ensures that clients learn consistent and compatible decision boundaries.

\begin{table}[!t]
\centering
\caption{Ablation study of the temporal modeling component on UCF-Crime.}
\label{tab:ablation_temporal}
\resizebox{0.65\columnwidth}{!}{
\begin{tabular}{l|c|c}
\toprule
\textbf{Temporal Module} & \textbf{AUC} & \textbf{mAUC} \\
\midrule
Frame-wise Similarity & 74.08 & 83.53 \\
3-Layer MLP & 84.44 & 88.10 \\
Axial Transformer & \textbf{86.36} & \textbf{91.06} \\
\bottomrule
\end{tabular}
}
\end{table}

\subsubsection{Impact of Temporal Modeling}
To isolate the contribution of the temporal module $\mathcal{T}$ in FedVAR, we compare the proposed structured attention mechanism against two structural variants. First, we remove the temporal module entirely by deriving anomaly scores directly from the frame-level vision–language similarities, thereby eliminating any learned temporal dependency. Second, we replace the transformer with a 3-Layer MLP operating over temporally flattened features. This preserves the overall representational capacity and allows for limited cross-frame interaction without employing a structured attention mechanism. The experimental results are presented in Table~\ref{tab:ablation_temporal}.

The results clearly demonstrate that explicit temporal reasoning is critical for video anomaly recognition. The frame-wise similarity variant achieves the lowest performance (74.08 AUC), indicating that relying solely on independent visual-textual alignment is insufficient for distinguishing complex anomalies that unfold over time. While introducing the 3-layer MLP provides a substantial improvement (+10.36\% AUC) by enabling basic cross-frame interactions, it still falls short of the full model. The proposed Axial Transformer achieves the highest performance (86.36 AUC and 91.06 mAUC). This confirms that simple non-linear transformations over flattened features cannot adequately substitute the explicit, structured attention mechanisms required to capture long-range temporal dependencies and refine anomaly localization across video sequences.

\subsubsection{Impact of Prompt Learning Strategies}
To evaluate the effectiveness of prompt learning in guiding the model through the pre-trained CLIP feature space, we ablate the design of the textual context vectors. Specifically, we compare our learnable prompt approach against a baseline using fixed, hand-crafted prompts (e.g., ``This is a video of a \{class\}''). Furthermore, to justify the parameterization of our context vectors, we compare a shared context token design (where a single set of learnable tokens is shared across all anomaly classes) against our default class-specific context token approach (where each anomaly class possesses unique learnable tokens). The experimental results on the UCF-Crime dataset are detailed in Table~\ref{tab:ablation_prompt}.

\begin{table}[!t]
\centering
\caption{Ablation study of textual prompt strategies on UCF-Crime.}
\label{tab:ablation_prompt}
\resizebox{0.70\columnwidth}{!}{
\begin{tabular}{l|c|c}
\toprule
\textbf{Prompt Strategy} & \textbf{AUC} & \textbf{mAUC} \\
\midrule
Manual Prompts & 84.94 & 85.31 \\
Shared Context $t^{ctx}$ & 85.85 & 90.16 \\
Class-Specific Context $t^{ctx}$ & \textbf{86.36} & \textbf{91.06} \\
\bottomrule
\end{tabular}
}
\end{table}

The results indicate that navigating the CLIP space benefits from optimal textual representations. The manual prompts yield the lowest performance (85.31 mAUC), demonstrating that static, engineered templates are insufficient for capturing the complex and highly variable visual semantics of video anomalies.
Introducing learnable prompts provides an immediate performance boost. The shared contexts variant significantly improves the mAUC to 90.16\%, proving that continuous prompt tuning effectively adapts the CLIP space for domain-specific anomaly recognition. However, the class-specific context token achieves the best overall performance (86.36 AUC and 91.06 mAUC). This confirms that assigning unique learnable directions for each anomaly category provides the necessary fine-grained semantic anchors to robustly resolve client-side misalignment across highly diverse abnormal events.

\begin{table}[!t]
\centering
\caption{Comparison of different federated aggregation strategies on UCF-Crime.}
\label{tab:ablation_strategies}
\resizebox{0.6\columnwidth}{!}{
\begin{tabular}{l|c|c}
\toprule
\textbf{FL Strategy} & \textbf{AUC} & \textbf{mAUC} \\
\midrule
FedYogi~\cite{reddi2020adaptive}    & 84.58 & 86.18 \\
FedMedian~\cite{yin2018byzantine}  & 85.76 & 90.35 \\
FedAvgM~\cite{hsu2019measuring}    & 86.11 & 90.49 \\
FedAvg~\cite{mcmahan2017communication} & \textbf{86.36} & \textbf{91.06} \\
\bottomrule
\end{tabular}
}
\end{table}

\subsubsection{Impact of Federated Aggregation Strategies}
To evaluate the impact of different federated aggregation strategies on FedVAR performance, we benchmark our framework against four representative aggregation strategies: FedAvg~\cite{mcmahan2017communication}, FedAvgM~\cite{hsu2019measuring} (momentum-based), FedMedian~\cite{yin2018byzantine} (robust aggregation), and FedYogi~\cite{reddi2020adaptive} (adaptive optimization). For each strategy, we use the default hyperparameters originally proposed for each algorithm to ensure a standardized comparison without strategy-specific tuning. The results on the UCF-Crime dataset are summarized in Table~\ref{tab:ablation_strategies}.

The results demonstrate that FedVAR is highly compatible with coordinate-wise averaging and robust aggregation. 
FedYogi achieves competitive performance (84.58 AUC and 86.18 mAUC), indicating that adaptive optimization remains applicable within the FedVAR framework. 
FedMedian further confirms that our alignment mechanism produces a strong global consensus, making the model resilient even to non-weighted robust aggregation.
FedAvgM yields the second-highest overall AUC (86.11), indicating that server-side momentum effectively dampens the stochastic noise and stabilizes the global model trajectory in the presence of diverse client-side anomaly direction vectors.
FedAvg achieves the highest mAUC (91.06), suggesting that the prototype-aligned feature space is inherently well-conditioned for simple averaging across heterogeneous classes. 

Ultimately, the consistent performance across FedYogi, FedMedian, FedAvgM, and FedAvg highlights that the prototype-alignment mechanism, rather than the specific aggregation math, is the primary driver of FedVAR’s robustness in decentralized environments.

\subsubsection{Impact of Client Participation Ratio}\label{subsec:ablation_cpr}
To investigate the scalability and communication efficiency of FedVAR for real-world deployments, we analyze the impact of partial client participation on model performance. Using a total pool of $N=10$ clients under the Random Split on UCF-Crime, we vary the Client Participation Ratio (CPR) $ \in \{0.2, 0.4, 0.6, 0.8, 1.0\}$, representing the fraction of clients randomly sampled to perform local updates in each communication round. This experiment assesses the model's sensitivity to reduced data updates and identifies potential performance saturation points. The results are detailed in Table~\ref{tab:ablation_cpr}.

\begin{table}[!t]
\centering
\caption{Impact of CPR on UCF-Crime performance with N = 10 total clients.}
\label{tab:ablation_cpr}
\resizebox{0.4\columnwidth}{!}{
\begin{tabular}{l|c|c}
\toprule
\textbf{CPR} & \textbf{AUC} & \textbf{mAUC} \\
\midrule
0.2 & 85.60 & 89.98 \\
0.4 & 85.92 & 90.00 \\
0.6 & 85.98 & 90.30 \\
0.8 & 86.05 & 90.42 \\
1.0 & \textbf{86.36} & \textbf{91.06} \\
\bottomrule
\end{tabular}
}
\end{table}

The empirical results reveal that FedVAR exhibits stable performance across varying participation regimes. Even with only 20\% client participation, the framework achieves 85.60 AUC, a marginal degradation of only 0.76\% compared to full participation. This suggests that the global normality prototype can be effectively maintained even with sparse updates, as the shared semantic anchor provides a strong inductive bias that remains consistent across partial client subsets. For real-world deployment, these findings imply that FedVAR can maintain high-fidelity anomaly recognition even under high client volatility or limited communication bandwidth, as full participation is not strictly required to achieve near-optimal performance.

It is also important to clarify the relationship between partial client participation and the global normality prototype $m_g$. As detailed in Section~\ref{sec:fedvar_framework}, the global prototype $m_g$ is aggregated only once prior to the start of the iterative federated training rounds. For the results presented in Table~\ref{tab:ablation_cpr}, $m_g$ was computed using the initial pool of available participating clients before the FL rounds began. In a real-world dynamic deployment where edge clients frequently drop out or join, newly joined or previously absent clients do not need to retrain the prototype. Instead, they can simply download the current $m_g$ from the server to immediately align their local feature spaces. To account for long-term distributional shifts in the edge network, the server can periodically update $m_g$ (e.g., via an exponential moving average of newly uploaded local prototypes from joining clients) without requiring a complete restart of the federated training process.

\subsubsection{Robustness to Client-Side Data Quality}\label{subsec:noisy_client}
To evaluate the resilience of FedVAR against low-quality or noisy decentralized client data, we introduce controlled label noise into a subset of clients. We vary the fraction of noisy clients ($\rho \in \{0.2, 0.4, 0.6\}$) and the local noise ratio ($\phi \in \{5\%, 10\%, 20\%, 50\%\}$), where $\phi$ represents the percentage of swapped normal and abnormal frame features. This setup simulates scenarios in which clients provide mislabeled or skewed samples for calculating the global normality prototype. The results are detailed in Table~\ref{tab:ablation_noise}.

\begin{table}[!t]
\centering
\caption{Robustness of FedVAR under varying fractions of noisy clients ($\rho$) and local noise ratios ($\phi$) on UCF-Crime.}
\label{tab:ablation_noise}
\resizebox{\columnwidth}{!}{
\begin{tabular}{c|c|c|c|c|c|c}
\toprule
\textbf{Noisy Clients ($\rho$)} & \multicolumn{2}{c|}{\textbf{20\%}} & \multicolumn{2}{c|}{\textbf{40\%}} & \multicolumn{2}{c}{\textbf{60\%}} \\ 
\cmidrule(lr){2-3} \cmidrule(lr){4-5} \cmidrule(lr){6-7}
\textbf{Noise Ratio ($\phi$)} & \textbf{AUC} & \textbf{mAUC} & \textbf{AUC} & \textbf{mAUC} & \textbf{AUC} & \textbf{mAUC} \\ 
\midrule
5\%  & 85.92 & 90.24 & 85.72 & 89.87 & 86.06 & 90.60 \\
10\% & 86.33 & 90.98 & 85.30 & 90.19 & 86.80 & 90.80 \\
20\% & 85.91 & 90.44 & 84.75 & 89.36 & 84.41 & 88.94 \\
50\% & 86.18 & 90.45 & 83.51 & 88.45 & 80.35 & 83.66 \\
\bottomrule
\end{tabular}
}
\end{table}

The empirical results demonstrate that FedVAR is highly resilient to moderate noise levels. When 20\% of the clients are noisy, the performance remains stable even at a 50\% noise ratio, with AUC and mAUC values (86.18 and 90.45) nearly identical to the clean baseline. This suggests that the weighted aggregation of local normality prototypes effectively filters out sparse semantic noise. 

As the fraction of noisy clients increases to 40\% and 60\%, performance remains robust for low-to-moderate noise ratios ($\phi \leq 10\%$). Significant degradation is only observed in the extreme case where 60\% of the network provides 50\% corrupted data (80.35 AUC), as the global normality prototype begins to be dominated by the misaligned representations. These findings confirm that FedVAR’s prototype-alignment mechanism provides a robust semantic anchor for the decentralized network, ensuring consistent recognition performance even when a significant portion of edge clients possess noisy data.

\subsection{System Efficiency and Overhead Analysis}
\label{sec:system_efficiency}
To complement the theoretical analysis and validate the feasibility of FedVAR for real-world edge deployment, we empirically evaluate the system overhead compared to state-of-the-art baselines. We measure three critical metrics: Trainable Parameters (in Millions, M), Computational Cost per video sequence (in Giga Floating-Point Operations, GFLOPs), and the Communication Cost per federated round (in Megabytes/Round, MB/Rd). The results are summarized in Table~\ref{tab:complexity}.

\begin{figure*}[!t]
    \centering
    \includegraphics[width=1.0\textwidth]{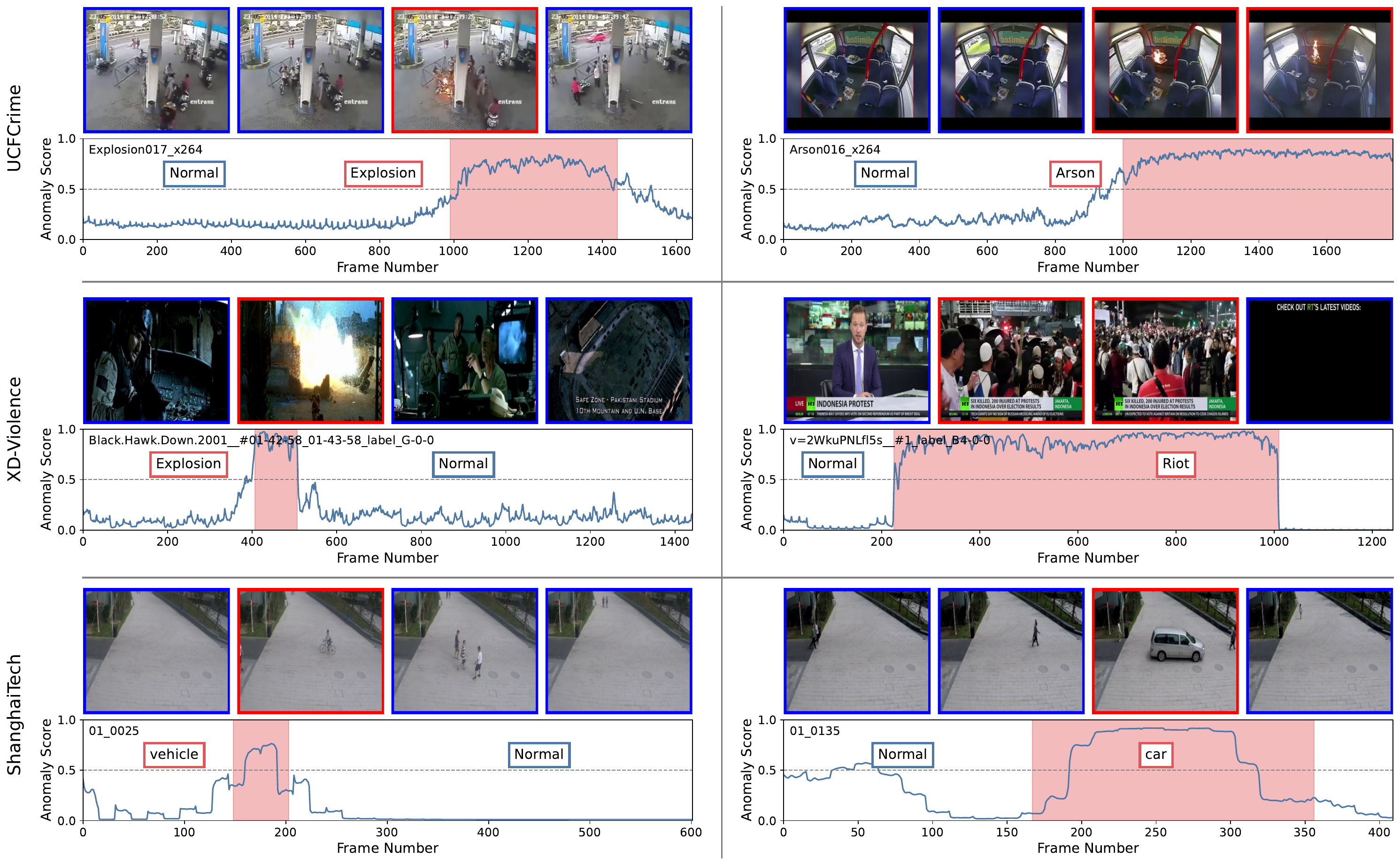}
    \caption{Qualitative visualization of anomaly predictions from FedVAR on representative test videos from UCF-Crime, XD-Violence, and ShanghaiTech. For each video, the lower plots depict the predicted frame-level anomaly probabilities and predicted anomaly classes over time, while the red shaded regions indicate ground-truth anomalous intervals. Example frames above the plots show red boxes marking detected anomalous segments and blue boxes denoting normal segments. The close correspondence between predicted scores and ground-truth intervals demonstrates FedVAR’s ability to maintain temporal consistency and robust anomaly recognition across domains.}
    \label{fig:var_qualitative}
\end{figure*}

As shown in Table~\ref{tab:complexity}, ZS-CLIP requires zero trainable parameters and no communication overhead. However, relying entirely on dense and unoptimized frame-text similarities results in the highest computational cost (54.25 GFLOPs) and lacks the necessary temporal reasoning for video anomaly recognition. FedCoOp exhibits the lowest communication overhead (0.56 MB/Rd) by exclusively learning textual prompts, but as demonstrated in previous sections, this comes at the cost of significantly lower recognition accuracy. 

In contrast, FedVAR strikes an optimal balance. By keeping the heavy CLIP visual and textual backbones completely frozen, it requires only 10.43M trainable parameters, comprising the prompt context tokens and the lightweight Axial Transformer. This results in a highly efficient communication cost of 79.57 MB per round. It is worth noting that our proposed global normality prototype alignment adds virtually zero communication overhead, as it only requires uploading a single $D$-dimensional vector from each client exactly once prior to the start of the FL rounds.

Computationally, because spatial features can be cached, the local computational cost is restricted to 12.03 GFLOPs per sequence. Compared to the most recent state-of-the-art baseline, Fed-WSVAD, FedVAR requires slightly less communication bandwidth (79.57 MB vs. 82.60 MB). While our framework imposes a marginally higher computational cost (12.03 vs. 7.70 GFLOPs), this is due to the structured temporal interactions of the axial attention mechanism. Notably, the Axial Transformer scales at $\mathcal{O}(T\sqrt{T})$ with respect to sequence length, making it fundamentally more efficient than standard $\mathcal{O}(T^2)$ temporal self-attention used in standard architectures, allowing it to process long untrimmed videos efficiently.

Ultimately, these metrics serve as hardware-agnostic indicators of efficiency. Because our empirical evaluation was conducted on a single GPU (NVIDIA A100) using pre-extracted CLIP visual features, these figures do not fully encapsulate real-world edge constraints such as strict memory bottlenecks, continuous I/O latency, or thermal limits. Nonetheless, compared to existing baselines, FedVAR demonstrates a highly favorable efficiency profile, making it a strong and scalable candidate for deployment in decentralized IIoT and CPS environments.

\begin{table}[!t]
\centering
\caption{Comparison of system efficiency and overhead. GFLOPs are calculated per video sequence. Communication cost represents the total upload and download size per client during a single federated communication round.}
\label{tab:complexity}
\resizebox{0.95\columnwidth}{!}{
\begin{tabular}{l|c|c|c}
\toprule
\textbf{Method} & \textbf{Params (M)} & \textbf{GFLOPs} & \textbf{Comm (MB/Rd)} \\
\midrule
ZS-CLIP \cite{zsclip_icml2021} & 0.00 & 54.25 & 0.00 \\
Temp-CLIP \cite{temporalclip_eccv2022} & 3.48 & 5.05 & 26.61 \\
FedCoOp \cite{fedcoop_tmc2023} & 0.07 & 2.17 & 0.56 \\
Fed-WSVAD \cite{fedwsvad_aaai2025} & 10.82 & 7.70 & 82.60 \\
\midrule
\textbf{FedVAR (ours)} & \textbf{10.43} & \textbf{12.03} & \textbf{79.57} \\
\bottomrule
\end{tabular}
}
\end{table}

\subsection{Qualitative Results}
To further demonstrate the interpretability and robustness of FedVAR, we visualize the temporal prediction patterns across representative test videos from UCF-Crime, XD-Violence, and ShanghaiTech. Figure~\ref{fig:var_qualitative} presents qualitative examples showing the frame-level anomaly probabilities and corresponding predicted anomaly classes produced by our model. As observed, FedVAR generates temporally consistent anomaly scores that closely follow the ground-truth intervals. The predicted abnormal frames typically coincide with key abnormal moments such as explosions, physical assaults, or vehicles intruding into pedestrian areas, while normal scenes maintain consistently low anomaly probabilities. Notably, the anomaly scores exhibit smooth temporal transitions rather than abrupt fluctuations, reflecting the stability introduced by prototype-guided alignment and indicating that the model has learned a coherent representation of abnormal patterns across videos. Overall, these qualitative observations align with the quantitative trends reported earlier, confirming that the proposed federated prompt aggregation enables consistent anomaly localization and recognition under heterogeneous and unseen conditions.

\section{Discussion and Limitations}
\label{sec:discussion}
While FedVAR demonstrates strong capabilities in establishing semantic consistency across decentralized clients, it is equally important to critically analyze its performance boundaries. In this section, we discuss the current limitations of our framework, contextualize these constraints within practical deployment scenarios, and outline concrete directions for future improvement.

First, despite its robust semantic alignment, FedVAR exhibits a slight performance gap compared to Fed-WSVAD on the XD-Violence benchmark (mAP of 47.93 vs. 49.33, as shown in Section \ref{subsec:var_results}). We attribute this to the inherent structural differences in handling extreme temporal sparsity. XD-Violence consists of long, untrimmed videos where anomalous segments are exceedingly short. Fed-WSVAD utilizes dynamic, context-aware prompt generation specifically optimized to capture and localize transient anomaly peaks in such long sequences. In contrast, FedVAR prioritizes semantic class separation (recognition) across clients via its prototype anchor, while relying on a standard Axial Transformer for temporal modeling. In extremely sparse videos, this coarse temporal aggregation can slightly dilute the anomaly localization signal. To close this gap, future iterations of FedVAR could integrate fine-grained, multi-scale temporal modeling or temporal contrastive learning to enhance boundary localization without sacrificing semantic consistency.

A second limitation pertains to the aggregation of the global normality prototype. Currently, FedVAR utilizes a sample-size weighted average (Eq.~\ref{eq:global_norm_proto}), which implicitly assumes that the quantity of normal frames is a reliable proxy for data quality. However, in practical Industrial IoT and Cyber-Physical Systems, data quality is often compromised by sensor degradation, environmental noise, or highly skewed local distributions. As demonstrated in our noise injection experiments (Section~\ref{subsec:noisy_client}), FedVAR demonstrates resilience to sparse semantic noise in client data. However, the semantic integrity of the global normality prototype can degrade under extreme conditions, as it becomes dominated by misaligned representations. To improve the framework's robustness against such highly degraded edge environments, future work could explore robust statistical aggregation strategies. Replacing the simple weighted average with a Coordinate-wise Median or a Trimmed Mean~\cite{yin2018byzantine} would allow the server to discard extreme ``semantic outliers'' before calculating the global prototype. Furthermore, integrating Byzantine-robust aggregators, such as Krum~\cite{allouah2024byzantine}, could secure the global semantic anchor against both unintentional label noise and targeted adversarial poisoning, ultimately ensuring highly reliable decentralized intelligence.

Finally, while Section~\ref{sec:system_efficiency} highlights the relative efficiency of FedVAR, our current system overhead analysis is based on experiments conducted on a data-center GPU with pre-cached features. Real-world edge computing, IIoT, and CPS scenarios impose strict constraints on memory footprint, continuous I/O latency, and thermal limits. Comprehensive benchmarking on actual edge hardware (e.g., NVIDIA Jetson platforms) and integrating edge-specific optimizations, such as feature quantization, adaptive caching, or knowledge distillation, remain important directions for future work to validate the end-to-end operational viability of the framework on resource-constrained devices.

\section{Conclusion}
In this paper, we introduced FedVAR, the first framework specifically designed to tackle the challenging task of fine-grained VAR within a privacy-preserving federated learning setting. We identified semantic misalignment, the divergence of client feature spaces due to highly non-IID data, as the critical barrier to effective federated VAR. To overcome this, we propose a novel prototype-based alignment mechanism. By aggregating clients' local normality prototypes into a single global semantic anchor, all clients explicitly re-center their visual and textual feature spaces. This technique enforces a consistent understanding of ``normality'' across the federation, enabling the learning of a robust and generalizable global model suitable for diverse edge environments.
Extensive experiments on three challenging benchmarks (UCF-Crime, XD-Violence, and ShanghaiTech) empirically validate the effectiveness of our approach. FedVAR consistently outperforms state-of-the-art federated baselines across diverse data partitioning schemes, demonstrating strong robustness to the client heterogeneity typical of real-world IIoT networks. Furthermore, we demonstrate strong generalization capabilities in both cross-domain settings and unseen anomaly classes. Overall, FedVAR successfully establishes a strong new baseline for decentralized video anomaly recognition. We believe this work paves the way for more practical, privacy-aware, and intelligent video analysis systems, contributing to the development of resilient cyber-physical infrastructures.

% =======================================================================================
% ========================  Appendix and Bibliography ===================================
% =======================================================================================
\section*{CRediT authorship contribution statement}
\textbf{Ghani Haider:} Conceptualization, Methodology, Software, Investigation, Writing – original draft, Writing – Reviewing and Editing, Visualization, Validation.
\textbf{Majid Kundroo:} Writing – Reviewing and Editing, Validation, Supervision.
\textbf{Boyun Eom:} Writing - Reviewing and Editing, Supervision.
\textbf{Dong-Hwan Park:} Writing- Reviewing and Editing, Supervision, Project administration, Funding acquisition.
\textbf{Chen Chen:} Writing – Reviewing and Editing, Validation.
\textbf{Taehong Kim:} Writing- Reviewing and Editing, Resources, Supervision, Project administration, Funding acquisition.

\section*{Declaration of competing interest}
The authors declare that they have no known competing financial interests or personal relationships that could have appeared to influence the work reported in this paper.

\section*{Data availability}
The datasets we used are publicly available.

\section*{Acknowledgments}
This work is supported by the Korea Agency for Infrastructure Technology Advancement (KAIA) grant funded by the Ministry of Land, Infrastructure and Transport (Grant: RS-2022-00155803).

\bibliographystyle{elsarticle-num} 
\bibliography{references}

\end{document}